\documentclass[10pt,journal,compsoc]{IEEEtran}
\ifCLASSOPTIONcompsoc
  \usepackage[nocompress]{cite}
\else
  \usepackage{cite}
\fi
\ifCLASSINFOpdf
  \usepackage[pdftex]{graphicx}
\else
\fi

\usepackage{enumitem}

\usepackage{hyperref} 
\hypersetup{
    colorlinks=true,
    linkcolor=red,
}
\usepackage{amsthm}

\usepackage{float}
\usepackage{arydshln}  
\usepackage{booktabs}
\usepackage{multirow}

\usepackage{caption}
\usepackage{subcaption}
\usepackage{amssymb}

\usepackage[table, svgnames, dvipsnames]{xcolor}

\begin{document}
%

\title{A Survey of Label-Efficient Deep Learning\\for 3D Point Clouds}
%
%
%
%

\author{Aoran~Xiao, Xiaoqin Zhang, Ling Shao~\IEEEmembership{Fellow,~IEEE},
        and~Shijian~Lu
\IEEEcompsocitemizethanks{
\IEEEcompsocthanksitem Aoran Xiao and Shijian Lu are with School of Computer Science and Engineering, Nanyang Technological University, Singapore.
\IEEEcompsocthanksitem Xiaoqin Zhang is with Key Laboratory of Intelligent Informatics for Safety \& Emergency of Zhejiang Province, Wenzhou University, China.
\IEEEcompsocthanksitem Ling Shao is with the UCAS-Terminus AI Lab, University of Chinese Academy of Sciences, Beijing, China.
}
}

%
%

\markboth{IEEE Transactions on Pattern Analysis and Machine Intelligence}
{Shell \MakeLowercase{\textit{et al.}}: Bare Demo of IEEEtran.cls for Computer Society Journals}
%



\IEEEtitleabstractindextext{%
\begin{abstract}
In the past decade, deep neural networks have achieved significant progress in point cloud learning. However, collecting large-scale precisely-annotated point clouds is extremely laborious and expensive, which hinders the scalability of existing point cloud datasets and poses a bottleneck for efficient exploration of point cloud data in various tasks and applications. Label-efficient learning offers a promising solution by enabling effective deep network training with much-reduced annotation efforts. This paper presents the first comprehensive survey of label-efficient learning of point clouds. We address three critical questions in this emerging research field: i) the importance and urgency of label-efficient learning in point cloud processing, ii) the subfields it encompasses, and iii) the progress achieved in this area. To this end, we propose a taxonomy that organizes label-efficient learning methods based on the data prerequisites provided by different types of labels. We categorize four typical label-efficient learning approaches that significantly reduce point cloud annotation efforts: data augmentation, domain transfer learning, weakly-supervised learning, and pretrained foundation models. For each approach, we outline the problem setup and provide an extensive literature review that showcases relevant progress and challenges. Finally, we share our views on the current research challenges and potential future directions. 
A project associated with this survey has been built at \url{https://github.com/xiaoaoran/3D_label_efficient_learning}. 
\end{abstract}

}

\maketitle

\IEEEdisplaynontitleabstractindextext

%
\IEEEpeerreviewmaketitle

\IEEEraisesectionheading{\section{Introduction}\label{sec:introduction}}
The acquisition of 3D point clouds has recently become more feasible and cost-effective with the wide adoption of various 3D devices, such as RGB-D cameras and LiDAR sensors. Meanwhile, remarkable advancements in deep learning have led to significant progress in point cloud understanding. The concurrence of the two has witnessed increasing demands in utilizing point clouds to capture 3D shape representations of objects and scenes, ranging from autonomous navigation to robotics and beyond.

Despite the great advancements in deep learning in point cloud understanding, most existing work relies heavily on large-scale well-annotated 3D data in network training. However, collecting such annotated training data is notoriously laborious and time-consuming due to the high complexity of the data, large variation in point sparsity, rich noises, and frequent 3D view changes in annotation process. Hence, how to learn effective point cloud models from training data of limited size and variation has become a grand challenge in point cloud understanding.

\begin{figure}[ht]
    \centering
    \includegraphics[width=\linewidth]{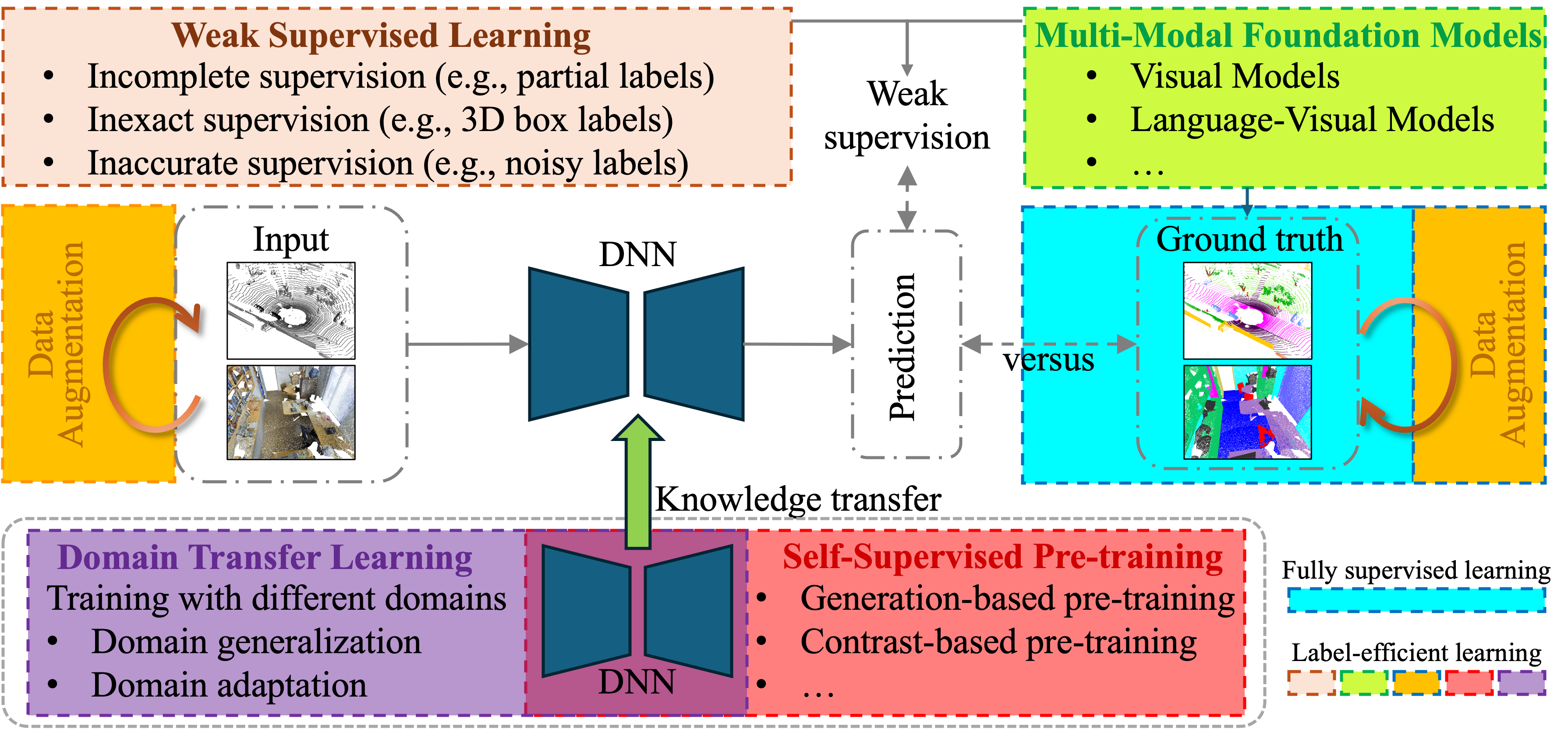}
    \vspace{-15pt}
    \caption{An overview of label-efficient learning for 3D point clouds. With the task of \textit{semantic segmentation}, we compare traditional \textcolor{cyan}{fully supervised learning}, which demands costly point-wise annotations, with label-efficient learning strategies prioritizing minimal annotation efforts. These strategies encompass \textcolor{YellowOrange}{data augmentation}, \textcolor{RawSienna}{weakly supervised learning}, \textcolor{Purple}{domain transfer learning}, \textcolor{WildStrawberry}{self-supervised pre-training} and \textcolor{LimeGreen}{multi-modal foundation models}. Best viewed in color.}
    \vspace{-15pt}
    \label{fig:overall-pipeline}
\end{figure}

To address the heavy burden in point cloud annotation, a promising solution is label-efficient learning, a machine learning paradigm that prioritizes model training with minimal annotation while still achieving desired accuracy. Due to its importance and high practical values, label-efficient point cloud learning has recently emerged as a thriving research field with numerous studies for learning effective models from limited point annotations. Various approaches have been explored with different data requirements and application scenarios. To this end, a systematic survey is urgently needed to provide a comprehensive overview of this field, covering multiple learning approaches and setups over various tasks in an organized manner.

\begin{figure*}[t]
    \centering
    \includegraphics[width=\linewidth]{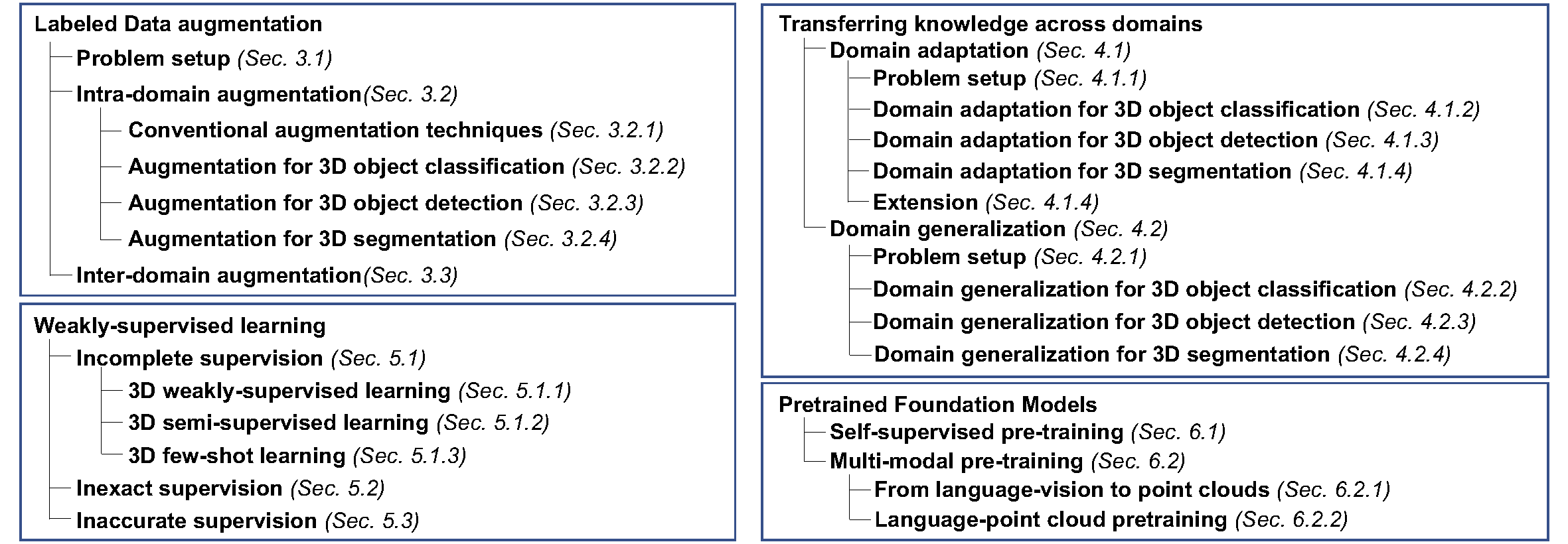}
    \vspace{-15pt}
    \caption{Taxonomy of label-efficient learning of point clouds.}
    \label{fig:taxonomy}
\end{figure*}

We thus present a comprehensive literature review of recent advancements in label-efficient learning of point clouds. Specifically, we review existing studies based on task and data prerequisites and categorize them into four distinct approaches: 1) \textit{Data Augmentation}, which expands limited labelled training data distribution via data augmentation; 2) \textit{Domain Transfer}, which utilizes labelled data from source domain(s) to train robust models for unlabelled target domain(s); 3) \textit{Weakly-Supervised Learning}, which trains robust models with weakly labelled point clouds; and 4) \textit{Pretrained Foundation Models}, which leverages unsupervised or multi-modal pretraining to facilitate 3D modelling with less annotations. 
Fig. \ref{fig:overall-pipeline} shows an overall picture of existing label-efficient learning approaches.
For each label-efficient learning approach, we introduce the problem setup and provide an exhaustive literature review, showcasing the progress made in this field and the challenges that remain.

To the best of our knowledge, this is the first systematic and comprehensive survey that focuses on label-efficient learning of point clouds, providing a detailed overview of the progress and challenges in this field. Several relevant surveys have been conducted. For example, \cite{guo2020deep,nguyen20133d} reviewed fully-supervised deep learning for various point clouds recognition tasks, and \cite{xiao2022unsupervised} presented a systematic review on unsupervised representation learning of point clouds. While these surveys provide valuable insights, they overlook label-efficient learning, a crucial task in many practical scenarios where labelled data is scarce.
In addition, several surveys~\cite{shen2023survey,jing2020self,qi2020small,wang2022generalizing,zhou2022domain} focus on label-efficient learning such as self-supervised learning~\cite{jing2020self,jaiswal2020survey,wilson2020survey}, small sample learning~\cite{qi2020small}, weakly-supervised learning\cite{zhou2018brief,van2020survey,ouali2020overview}, and generalizing across domains~\cite{wang2022generalizing,zhou2022domain}. However, they focus on different data modalities (e.g., 2D images, texts, and graphs) without covering the unique challenges and advancements in point cloud processing. We believe that this survey will serve as a valuable resource for researchers and practitioners alike by bridging the gap in the current literature and facilitating future development in this very promising field.

The rest of this survey is organized as follows. Sec.~\ref{sec:backgrounds} introduces background knowledge including key concepts and a brief description of the efforts and difficulty of annotating 3D point-cloud data. Sections~\ref{sec:data_aug},\ref{sec.domain_transfer},\ref{sec.weakly},\ref{sec.pretraining} then provide systematic and extensive literature reviews of four representative data-efficient learning approaches, namely, point cloud data augmentation, knowledge transfer across domains, weakly supervised learning of point clouds, and pretrained foundation models for point cloud learning. 
In Sec. \ref{sec.benchmarks}, we discuss pros and cons of each label-efficient learning approach, followed by a comprehensive benchmarking analysis across multiple public datasets, aiming to showcase strengths and weaknesses of previous endeavors in the field.
With the limitations and challenges of existing work, we identify and discuss several promising research directions for future label-efficient point cloud learning in Sec. \ref{sec.future}.
Fig.~\ref{fig:taxonomy} shows a taxonomy of existing label-efficient learning methods for 3D point clouds.

\section{Background} \label{sec:backgrounds}

This section covers the challenges of point cloud annotation and the significance of label-efficient learning for point clouds. For background information on point cloud deep learning and related tasks, please refer to the key concepts in the appendix.

\begin{table*}[t]
    \setlength{\tabcolsep}{1.6mm}
    \centering
    \scriptsize
    \vspace{-1mm}
    \caption{A summary of commonly used datasets for point cloud learning.}
    \vspace{-5pt}
    \begin{tabular}{|r|c|c|c|c|c|c|}
        \hline
        Dataset & Year & \#Samples & \#Classes & Type & Representation & Label\\
        \hline
        ModelNet40~\cite{wu20153d} & 2015 & 12,311 objects & 40 & Synthetic object & Mesh & Object category label\\
        ShapeNet~\cite{chang2015shapenet} & 2015 & 51,190 objects & 55 & Synthetic object & Mesh & Object/part category label\\
        ScanObjectNN~\cite{uy2019scanobjectnn} & 2019 & 2,902 objects & 15 & Real-world object & Points & Object category label\\
        SUN RGB-D~\cite{song2015sun} & 2015 & 5K frames & 37 & Indoor scene & RGB-D & 3D bounding box label\\
        S3DIS~\cite{armeni2016s3dis} & 2016 & 272 scans & 13 & Indoor scene & RGB-D & Point category label\\
        ScanNet~\cite{dai2017scannet} & 2017 & 1,513 scans & 20 & Indoor scene & RGB-D \& mesh & Point category label \& 3D bounding box label\\
        KITTI~\cite{geiger2013vision} & 2013 & 15K frames & 8 & Outdoor driving & RGB \& LiDAR & 3D bounding box label\\
        nuScenes~\cite{caesar2020nuscenes} & 2020 & 40K & 32 & Outdoor driving & RGB \& LiDAR & Point category label \& 3D bounding box label \\ 
        Waymo~\cite{sun2020scalability} & 2020 & 200K & 23 &  Outdoor driving & RGB \& LiDAR & Point category label \& 3D bounding box label \\ 
        STF~\cite{bijelic2020seeing} & 2020 & 13.5K & 4 &  Outdoor driving & RGB \& LiDAR \& Radar & 3D bounding box label\\
        ONCE~\cite{mao2021once} & 2021 & 1M scenes & 5 & Outdoor driving & RGB \& LiDAR & 3D bounding box label\\
        Semantic3D~\cite{hackel2017semantic3d} & 2017 & 15 dense scenes & 8 & Outdoor TLS & Points & Point category label\\ 
        SemanticKITTI~\cite{behley2019semantickitti} & 2019 & 43,552 scans & 28 & Outdoor driving & LiDAR & Point category label\\
        SensatUrban~\cite{hu2021towards} & 2020 & 1.2 $\mathrm{km}^2$  & 31 & UAV Photogrammetry & Points & Point category label\\
        SynLiDAR~\cite{xiao2022transfer}  & 2022 & 198,396 scans & 32 & Outdoor driving & Synthetic LiDAR & Point category label\\ 
        SemanticSTF~\cite{xiao20233d}  & 2023 & 2,086 scans & 21 & Outdoor driving & RGB \& LiDAR & Point category label\\ 
        \hline
    \end{tabular}
    \label{tab:datasets}
    \vspace{-2mm}
\end{table*}

\subsection{Annotation efforts for 3D datasets}\label{subsec:anno_efforts}
Annotating point clouds is challenging due to their unique data characteristics. Unlike images, point clouds are often incomplete, sparse, and lack color information, leading to ambiguities in both semantics and geometries. Additionally, changes in 3D views complicate the annotation process and can even cause motion sickness. Hence, 3D annotators require special training to produce accurate and consistent annotations for various 3D visual perception tasks.

\begin{table*}[h]
    \scriptsize
    \setlength\tabcolsep{3pt}
    \centering
    \caption{Summary of data augmentation methods. ``Cls", ``Det", and ``Seg" denote classification, detection, and segmentation, respectively.}
    \vspace{-5pt}
    \begin{tabular}{|l|c|c|c|l|}
    \hline
        Method & Published in & Domain & Task & Contribution \\
    \hline
        PointAugment~\cite{li2020pointaugment} & CVPR2020 & Intra-domain & Cls & Design an augmentor network to enhance classifier via adversarial learning.\\
        Pointmixup~\cite{chen2020pointmixup} & ECCV2020 & Intra-domain & Cls & Shortest path linear interpolation for mixing and generating augmented samples. \\
        RSMix~\cite{lee2021regularization} & CVPR2021 & Intra-domain & Cls & Replacing parts of samples for mixed virtual samples.\\
        PointWOLF~\cite{kim2021point} & ICCV 2021 & Intra-domain & Cls & Non-rigid deformations and AugTune for automated sample generation\\
        PointCutMix~\cite{zhang2022pointcutmix} & NeuroComputing2022 & Intra-domain & Cls & Replacing paired points from different objects by finding optimal assignments. \\
        Point MixSwap~\cite{umam2022point} & ECCV2022 & Intra-domain & Cls & Attention-based augmentation by swapping object subsets of the same class. \\
        SageMix~\cite{lee2022sagemix} & NeurIPS2022 & Intra-domain & Cls & Mixup objects by blending salient regions with re-weighted saliency scores.\\
        GT-Sampling~\cite{yan2018second} & sensors2018 & Intra-domain & Det & Copy instances from other point cloud scenes and paste them into the current one.\\
        PPBA~\cite{cheng2020improving} & ECCV2020 & Intra-domain & Det & Search for the automated data augmentation parameters by progressive population.\\
        AziNorm~\cite{chen2022azinorm} & CVPR2022 & Intra-domain & Det/Seg & Normalizes point clouds along the radial direction and eliminates the variability.\\
        Mix3D~\cite{nekrasov2021mix3d} & 3DV2021 & Intra-domain & Seg & Directly mix two scene-level point clouds and labels in an out-of-context way \\
        PolarMix~\cite{xiao2022polarmix} & NeurIPS2022 & Intra-domain & Det/Seg & Mix labelled LiDAR point clouds in the Polar System\\
    \hline
        PointAugmenting~\cite{wang2021pointaugmenting} & CVPR2021 & Inter-domain & Det & Expand GT-Sampling by pasting virtual objects into both images and point clouds. \\
        LiDAR-Aug~\cite{fang2021lidar} & CVPR2021 & Inter-domain & Det & Place CAD objects into LiDAR scans and render for real occlusion. \\
        PCT~\cite{xiao2022transfer} & AAAI2022 & Inter-domain & Seg & Combine synthetic and real LiDAR point clouds for joint training\\
    \hline
    \end{tabular}
    \label{tab. Sum of data augmentation methods}
\end{table*}

Meanwhile, fully automated point cloud annotation is still infeasible at this stage. Several semi-automatic tools have been developed to ease the problem, but they still require tedious manual effort to ensure accuracy and many of them do not generalize well.
For example, SemanticKITTI~\cite{behley2019semantickitti} uses dense point clouds from superimposed LiDAR scans; however, this superimposition requires precise localization and can distort moving objects. Consequently, point cloud annotation still involves manual effort heavily, demanding extensive time and effort from well-trained annotators.

The labor-intensive nature of point cloud annotation limits the size and diversity of public datasets, as shown in Table~\ref{tab:datasets}. This poses a significant challenge for developing generalizable learning algorithms. Therefore, label-efficient learning is essential and urgent to overcome the limitations of existing point cloud data.

\subsection{Related Applications}

Label-efficient learning has been extensively explored in various 3D tasks. For example, data augmentation, as reviewed in Sec. \ref{sec:data_aug}, has been widely adopted while training 3D models. In addition, it has shown that models trained with a tiny amount of human annotation, e.g., 0.1\% point labels, can produce similar segmentation as fully supervised models trained with 100\% point labels~\cite{liu2022less}.

Supporting software tools further facilitate label-efficient learning with point clouds. For example, NVIDIA Omniverse \cite{nvidiaomniverse} offers a suite of tools for 3D simulation, rendering, and collaboration, leading to synthetic, realistic, and self-labelled point clouds across scenarios. Similarly, Carla \cite{Dosovitskiy17} and AirSim \cite{airsim2017fsr} simulate LiDAR sensors in virtual driving scenes, eliminating the laborious point cloud collection and annotation process.

\section{Data Augmentation}\label{sec:data_aug}
\begin{figure}[t]
    \centering
    \includegraphics[width=0.9\linewidth]{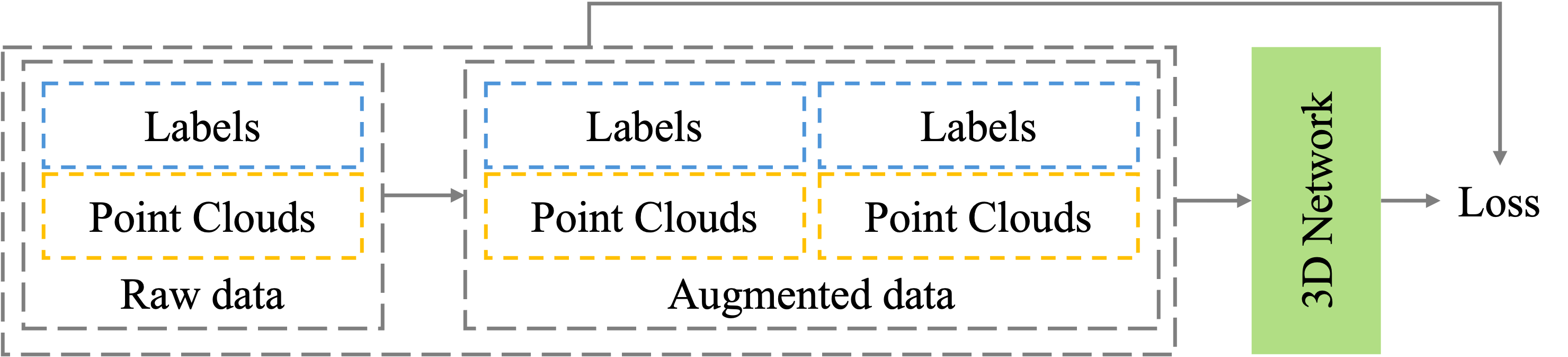}
    \vspace{-5pt}
    \caption{Data augmentation in 3D network training.}
    \label{fig:da}
    \vspace{-10pt}
\end{figure}

Data augmentation (DA), which aims to increase data size and data diversity by generating new training data from existing data as illustrated in Fig.~\ref{fig:da}, has been widely explored in deep network training~\cite{shorten2019survey}. It has demonstrated great effectiveness across various tasks, especially when only limited training data is available.

Most existing DA methods for point clouds can be broadly grouped into two categories: \textit{intra-domain} DA and \textit{inter-domain} DA. Intra-domain DA generates training data from existing ones (Sec. \ref{subsec.intra_da}), and it has been tackled via conventional augmentation techniques (Sec. \ref{sec.DA_CDA}), as well as specific techniques for 3D shape classification (Sec. \ref{sec.DA_cls}), object detection (Sec. \ref{sec.DA_det}), and segmentation (Sec. \ref{sec.DA_seg}). Inter-domain DA introduces other data sources, such as synthetic or cross-modal data, to expand the distribution of existing training data (Sec. \ref{subsec.inter_da}). Table~\ref{tab. Sum of data augmentation methods} shows representative DA methods.

\subsection{Intra-domain augmentation}\label{subsec.intra_da}

\subsubsection{Conventional augmentation techniques}\label{sec.DA_CDA}
\begin{figure}[t]
    \centering
    \includegraphics[width=\linewidth]{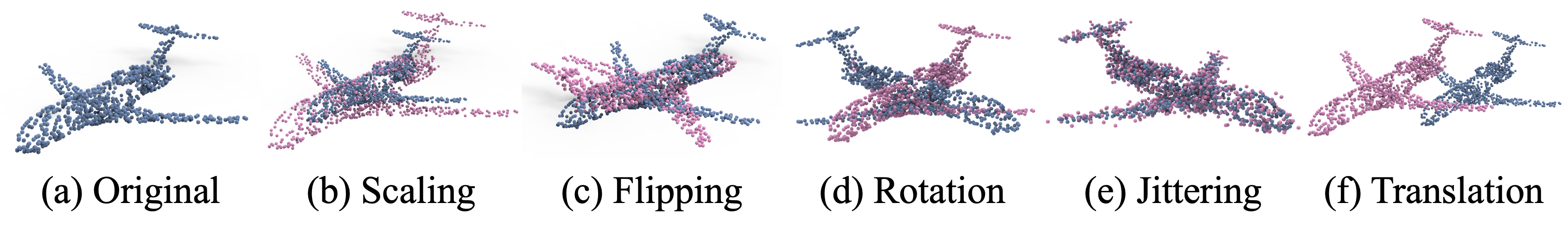}
    \vspace{-15pt}
    \caption{Illustration of widely-used conventional augmentation techniques for point clouds.}
    \label{fig:DA-CDA}
    \vspace{-3mm}
\end{figure}

Conventional DA has been extensively explored as a pre-processing operation in various 3D tasks~\cite{qi2017pointnet,qi2017pointnet++,qi2019deep,hu2020randla,xiao2021fps,xie2020pointcontrast}. It adopts different spatial transformations to generate diverse views of point clouds that are crucial for learning transformation-invariant and generalizable representations. Fig.~\ref{fig:DA-CDA} shows a list of typical conventional DA techniques together with qualitative illustrations.

\begin{itemize}[noitemsep,topsep=0pt]
    \item \textbf{Scaling} changes the scale of the point cloud by multiplying the coordinates with a ratio $s$, where a value of $s<1$ indicates shrinkage and $s>1$ indicates enlargement as illustrated in Fig.~\ref{fig:DA-CDA} (b). 
    \item \textbf{Flipping} randomly flips points along the x-axis or y-axis, as illustrated in Fig.~\ref{fig:DA-CDA} (c).
    \item \textbf{Rotation} rotates the points around the z-axis with a random angle, as illustrated in Fig.~\ref{fig:DA-CDA} (d).
    \item \textbf{Jittering} adds random perturbations to point clouds with Gaussian noise with zero mean and a standard deviation of $\beta$\cite{qi2017pointnet}, as illustrated in Fig.\ref{fig:DA-CDA} (e).
    \item \textbf{Translation} involves shifting all points in the same direction and distance, as shown in Fig.~\ref{fig:DA-CDA} (f).
\end{itemize}

\noindent Note conventional DA can be applied to both global point clouds and local point patches~\cite{hahner2020quantifying,sheshappanavar2021patchaugment,choi2021part}. Despite its simplicity and popularity, conventional DA often leads to insufficient network training. The major reason is that network training is often independent of DA process, offering little feedback for DA optimization. In addition, new training samples are often augmented from individual instead of multiple existing samples, leading to limited expansion of data distribution. This issue has been studied for various point cloud tasks, more details to be elaborated in subsequent subsections.

\vspace{-5pt}
\subsubsection{DA for 3D shape classification}\label{sec.DA_cls}
Adaptive DA has been explored for 3D shape classification. For example, Li et al.~\cite{li2020pointaugment} developed Pointaugment that generates training data with shape-wise transformations and point-wise displacements, optimized jointly with the classifier through adversarial learning. Kim et al.~\cite{kim2021point} designed AugTune that controls local deformations adaptively to generate realistic samples with large pose diversity and shape identity preservation.

\begin{figure}[t]
    \centering
    \includegraphics[width=\linewidth]{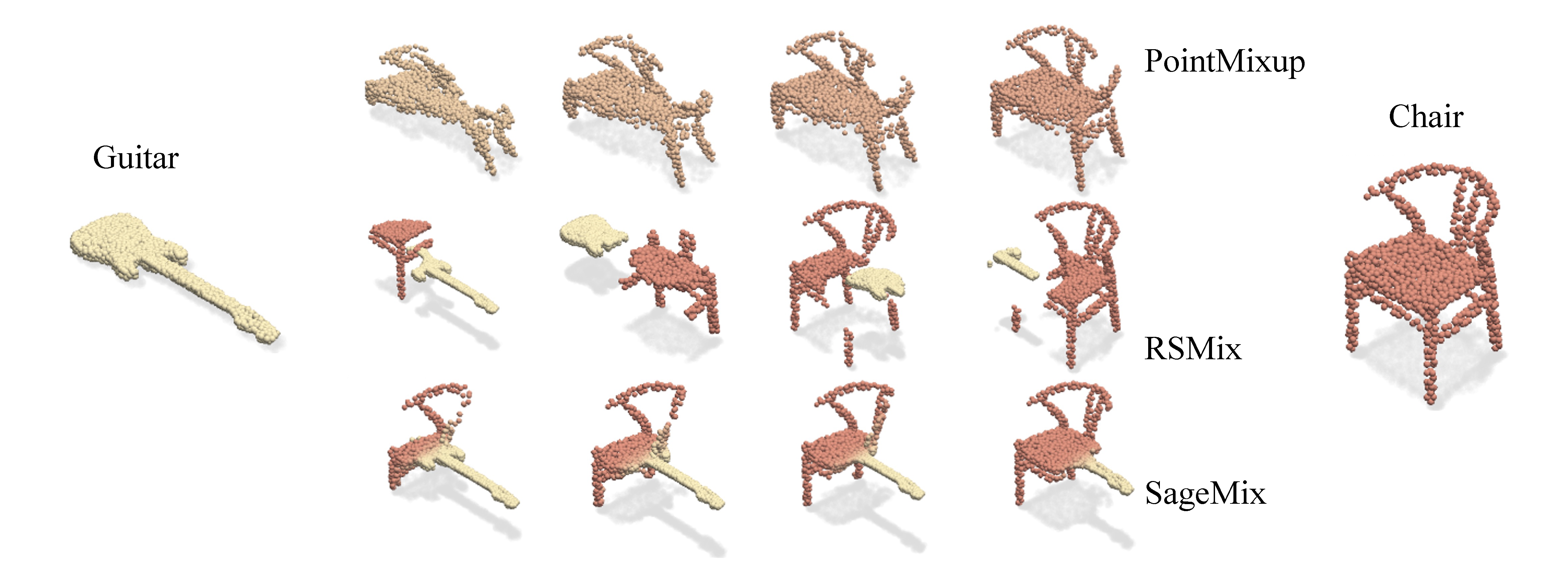}
    \vspace{-15pt}
    \caption{Illustration of typical mixing DA methods in point cloud classification, including PointMixup~\cite{chen2020pointmixup}, RSMix~\cite{lee2021regularization}, and SageMix~\cite{lee2022sagemix}.}
    \label{fig:DA-CLS}
    \vspace{-10pt}
\end{figure}

Another line of research mixes existing objects to create new training samples. Inspired by MixUp~\cite{zhang2017mixup,verma2019manifold}, \cite{chen2020pointmixup} presents PointMixup that interpolates objects of different classes via shortest path linear interpolation. Several approaches further explored how to preserve local object structures as illustrated in Fig.~\ref{fig:DA-CLS}, including subset mixup in RSMix~\cite{lee2021regularization}, local part replacement in PointCutMix \cite{zhang2022pointcutmix}, saliency-guided mixup in SageMix~\cite{lee2022sagemix}, and intra-category mixup in Point-MixSwap~\cite{umam2022point}.

\vspace{-5pt}
\subsubsection{DA for 3D object detection}\label{sec.DA_det} 
3D object detection works with scene-level point clouds that are very different from object-level point clouds. Specifically, scene-level point clouds have much more points, more diverse surroundings, larger density variation, and more noises or outliers. 
Several work explores DA under such challenging scenario. For example, \cite{cheng2020improving} searches for optimal DA strategies via progressive population-based augmentation. \cite{chen2022azinorm} performs azimuth-normalization to address azimuth variation in LiDAR point clouds. \cite{leng2022pseudoaugment} exploits pseudo labels of unlabelled point clouds for DA for effective point cloud learning.

The mixing approach has also been applied to 3D object detection. Yan et al. \cite{yan2018second} proposed GT-Aug, which copies objects from other LiDAR frames and pastes them randomly into the current frame. However, this method ignores the real-world relationships between objects. To address this, Sun et al. \cite{Sun_Fang_Zhu_Li_Lu_2022} used a correlation energy field to represent these relationships during pasting. Additionally, Wu et al.~\cite{zheng2022boosting}  fused multiple LiDAR frames to create denser point clouds, enhancing object detection in single-frame scenarios.

\vspace{-5pt}
\subsubsection{DA for 3D semantic segmentation}\label{sec.DA_seg}

\begin{figure}[t]
    \centering
    \includegraphics[width=\linewidth]{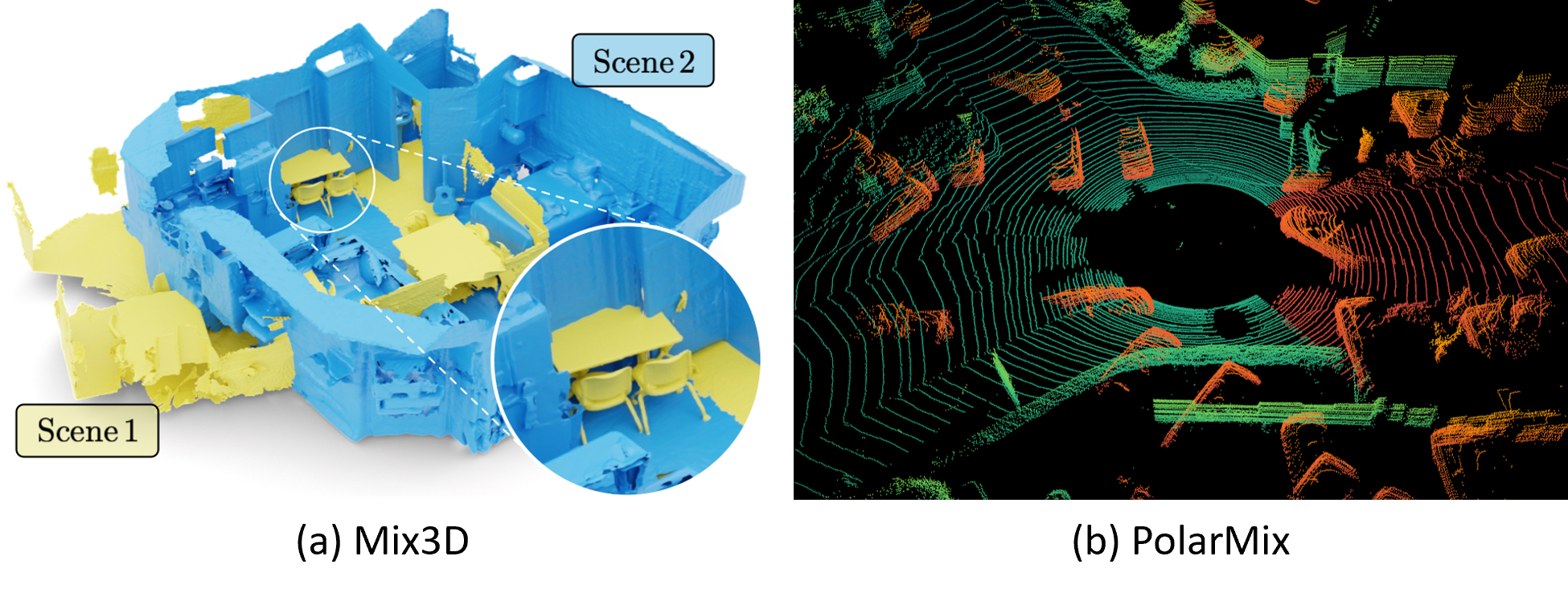}
    \vspace{-15pt}
    \caption{Mixing-based DA on point cloud semantic segmentation: Mix3D~\cite{nekrasov2021mix3d} performs out-of-context mixing, while PolarMix~\cite{xiao2022polarmix} applies in-context mixing. The two graphs are extracted from \cite{nekrasov2021mix3d} and \cite{xiao2022polarmix}.}
    \label{fig:DA-SemSeg}
    \vspace{-10pt}
\end{figure}

Mixing-based DA has significantly improved point cloud segmentation performance. Mix3D~\cite{nekrasov2021mix3d} concatenates two point clouds and their labels for out-of-context augmentation. PolarMix \cite{xiao2022polarmix} instead, mixes LiDAR frames in the polar coordinate system to maintain LiDAR-specific properties like partial visibility and density variation. They implemented scene-level swapping and instance-level rotate-pasting, achieving consistent augmentation across various LiDAR segmentation and detection benchmarks. Fig.~\ref{fig:DA-SemSeg} illustrates these methods qualitatively.

\subsection{Inter-domain augmentation}\label{subsec.inter_da}
Inter-domain DA enhances network training by using additional data including synthetic or cross-modality data.

\noindent\textbf{Synthetic data}. Fang et al. \cite{fang2021lidar}  introduced LiDAR-Aug, which inserts CAD objects into road scene point clouds to create richer training data for 3D detectors. Xiao et al. \cite{xiao2022transfer} used self-annotated LiDAR point clouds from game engines combined with real data to train 3D segmentation networks. However, the domain gap between synthetic and real point clouds can limit effectiveness \cite{xiao2022transfer}.

\noindent\textbf{Cross-modality data}. Several studies fuse point clouds with data of other modalities for alleviating the inherent limitations of 3D sensors. For example, RGB images are widely adopted to improve network training for 3D object detection~\cite{chen2017multi,qi2018frustum, liang2018deep,huang2020epnet,wang2021pointaugmenting,vora2020pointpainting} and 3D semantic segmentation~\cite{yan20222dpass}. Recently, several studies~\cite{yang2020radarnet,qian2021robust} fused radar point clouds and LiDAR point clouds for learning more robust and generalizable point cloud models.

\subsection{Summary} 

Recent studies, as we reviewed in this section, have shown that DA can reduce the need for extensive data collection and annotation while achieving comparable performance. However, DA for point cloud learning still remains far under-explored, especially compared to 2D computer vision and NLP. Existing studies are limited, particularly for scene-level point clouds crucial in practical applications. Additionally, most DA methods are tailored for specific datasets and may not generalize well to real-world point clouds with greater variation. More research is needed to advance this promising field.

\section{Domain Transfer Learning}\label{sec.domain_transfer}

Domain transfer learning leverages knowledge from previously collected and annotated data to handle new data, significantly reducing labelling efforts. However, transferring knowledge across different domains often faces \textit{domain discrepancy}~\cite{ben2010theory}, a distributional bias or shift between data from different domains. Consequently, models trained on source-domain data often perform poorly on target-domain data due to this discrepancy. This issue has been studied through two major setups: \textit{domain adaptation} and \textit{domain generalization}. Domain adaptation adapts a machine learning model trained on one domain to perform well on another by minimizing the distribution shift between the two domains. In contrast, domain generalization learns common and invariant features from the training domain(s), aiming to develop a model that performs well on new, unseen domains. While both approaches aim to create robust models that perform well on target data, domain adaptation allows access to target data during training, whereas domain generalization does not.

\subsection{Domain adaptation}\label{subsec.domain_adaptation}

\begin{table*}[h]
    \scriptsize
    \setlength\tabcolsep{1pt}
    \centering
    \caption{Summary of domain adaptation methods. $^{\dag}$ means UDA; $^{\ddag}$ represents test time adaptation; $^{\sharp}$ means source-free UDA; $^{\natural}$ denotes multi-modal learning.}
    \vspace{-5pt}
    \begin{tabular}{|l|c|l|l|}
    \hline
        Method & Published in & Task & Contribution \\
    \hline
        Pointdan~\cite{qin2019pointdan} & NeurIPS 2019 & Classification$^{\dag}$ & Jointly align the global and local features across domains of object-level point clouds in multi-level. \\
        DefRec~\cite{achituve2021self} & WACV 2021 & Classification$^{\dag}$ & Deformation Reconstruction and mixup of object-level point clouds for synthetic-to-real adaptation. \\
        GAST~\cite{zou2021geometry} & ICCV 2021 & Classification$^{\dag}$ & A geometry-aware self-training. \\
        GLRV~\cite{fan2022self} & CVPR 2022 & Classification$^{\dag}$ & Adaptation with modeling  global-local structures. \\
        IPCDA~\cite{shen2022domain} & CVPR 2022 & Classification$^{\dag}$ & Adaptation by employing a self-supervised task of learning geometry-aware implicits. \\
        MLSP~\cite{liang2022point} & ECCV 2022 & Classification$^{\dag}$ & Learning shared feature space across domains by predicting masked local structures. \\ 
        PC-Adapter~\cite{park2023pc} & ICCV 2023 & Classification$^{\dag}$ & Adaptation with shape-aware adapter and locality-aware adapter. \\ 
    \hline
        SN~\cite{wang2020train} & CVPR 2020 & Detection$^{\dag}$ & Statistic normalization in car sizes across geographic areas.\\
        CDN~\cite{su2020adapting} & ECCV 2020 & Detection$^{\dag}$ & Conditionally encode different domains into a shared latent space with the same domain attribute. \\
        ST3D~\cite{yang2021st3d} & CVPR 2021 & Detection$^{\dag}$ & Self-training with quality-aware pseudo labelling and curriculum-based data augmentation.\\
        SRDAN~\cite{zhang2021srdan} & CVPR 2021 & Detection$^{\dag}$ & Adaptation with scale-aware domain alignment and range-aware domain alignment. \\
        FogSim~\cite{hahner2021fog} & ICCV 2021 & Detection$^{\dag}$ & Simulate fog noise for LiDAR point clouds. \\
        MLC-Net~\cite{luo2021unsupervised} & ICCV 2021 & Detection$^{\dag}$ & Exploit point-, instance- and neural statistics-level consistency for cross-domain adaptation. \\
        SPG~\cite{xu2021spg} & ICCV 2021 & Detection$^{\dag}$ & Recover missing parts of the foreground objects for better detection. \\
        3D-CoCo~\cite{yihan2021learning} & NeurIPS 2021 & Detection$^{\dag}$ & Contrastive Co-training for BEV-based 3D adaptive detection.\\
        SnowSim~\cite{hahner2022lidar} & CVPR 2022 & Detection$^{\dag}$ & Simulate snow noise for LiDAR point clouds. \\
        LiDARDistill~\cite{wei2022lidar} & ECCV 2022 & Detection$^{\dag}$ & A progressive framework to mitigate the beam-induced domain shift. \\
        CL3D~\cite{} & AAAI 2022 & Detection$^{\dag}$ & Self-training with spatial geometry alignment and temporal motion alignment. \\
        DTS~\cite{hu2023density} & CVPR 2023 & Detection$^{\dag}$ & A density-insensitive domain adaption framework. \\
        GPA-3D~\cite{li2023gpa} & CVPR 2023 & Detection$^{\dag}$ & Geometry-aware prototype alignment for BEV representations across domains. \\
        Bi3D~\cite{yuan2023bi3d} & CVPR 2023 & Detection$^{\dag}$ & Active learning with domainness-aware source sampling and diversity-based target sampling.\\
        ReDB~\cite{chen2023revisiting} & ICCV 2023 & Detection$^{\dag}$ & A cross-domain mixing, density-invariant, and class-balanced self-training solution.\\
    \hline
        SqueezeSegV2~\cite{wu2019squeezesegv2} & ICRA 2019 & Semantic Segmentation$^{\dag}$ & Adapting in 2D projection space with intensity rendering, geodesic alignment, and normalization.\\
        Complete\&Label~\cite{yi2021complete} & CVPR 2021 & Semantic Segmentation$^{\dag}$ & Tackle domain adaptation by transforming it into a 3D surface completion task.\\ 
        ePointDA~\cite{zhao2021epointda} & AAAI 2021 & Semantic Segmentation$^{\dag}$ & 2D projection-based adaptation with noise rendering, statistic invariation, and spatially-adaption. \\
        PCT~\cite{xiao2022transfer} & AAAI 2022 & Semantic Segmentation$^{\dag}$ & Translation from synthetic to real LiDAR point clouds for domain adaptation.\\
        CoSMix~\cite{saltori2022cosmix} & ECCV 2022 & Semantic Segmentation$^{\dag}$ & Inter-domain mixing for 3D  adaptive segmentation.\\
        GIPSO~\cite{saltori2022gipso} & ECCV 2022 & Semantic Segmentation$^{\sharp}$ & Source-free online UDA with adaptive self-training and geometric-feature propagation. \\
        ASM~\cite{Li_2023_CVPR} & CVPR 2023 & Semantic Segmentation$^{\dag}$ & Adversarial training with learnable masking for modeling target noise.\\
        UniMix~\cite{zhao2024unimix} & CVPR 2024 & Semantic Segmentation$^{\dag}$ & Mixing across weather over spatial, intensity, and semantic distributions. \\
        DGT-ST~\cite{yuan2024density} & CVPR 2024 & Semantic Segmentation$^{\dag}$ & A density-guided translator with a two-stage self-training pipeline. \\
    \hline
        xMUDA~\cite{jaritz2020xmuda} & CVPR 2020 & Semantic Segmentation$^{\dag\natural}$ & Cross-modal UDA for 3D semantic segmentation with paired 2D images and 3D point clouds. \\
        DsCML~\cite{peng2021sparse} & ICCV 2021 & Semantic Segmentation$^{\dag\natural}$ & Dynamic sparse-to-dense and adversarial cross-modal learning for 3D adaptive segmentation. \\
        MM-TTA~\cite{shin2022mm} & CVPR 2022 & Semantic Segmentation$^{\ddag\natural}$ & Multi-modal test-time adaptation for 3D semantic segmentation. \\
        MM-CTTA~\cite{cao2023multi} & ICCV 2023 & Semantic Segmentation$^{\ddag\natural}$ & Multi-modal continual test-time adaptation for 3D semantic segmentation. \\
    \hline
    \end{tabular}
    \label{tab. Sum of domain adaptation methods}
    \vspace{-5pt}
\end{table*}

Domain adaptation provides an economical solution for utilizing existing annotated training data with the same label space for fine-tuning models from a source domain to a target domain. For point clouds, domain adaptation studies vary based on data prerequisites and application scenarios. Most existing research focuses on unsupervised domain adaptation (UDA), which learns from labelled source point clouds and unlabelled target point clouds. This section presents the problem setup of UDA in Section \ref{subsec.uda_set}, UDA for 3D shape classification in Section \ref{DAdapt-cls}, UDA for 3D object detection in Section \ref{DAdapt-det}, UDA for 3D segmentation in Section \ref{DAdapt-seg}, and other types of domain adaptation for point clouds in Section \ref{subsec.uda_extension}.

\begin{figure}[t]
    \centering
    \includegraphics[width=0.9\linewidth]{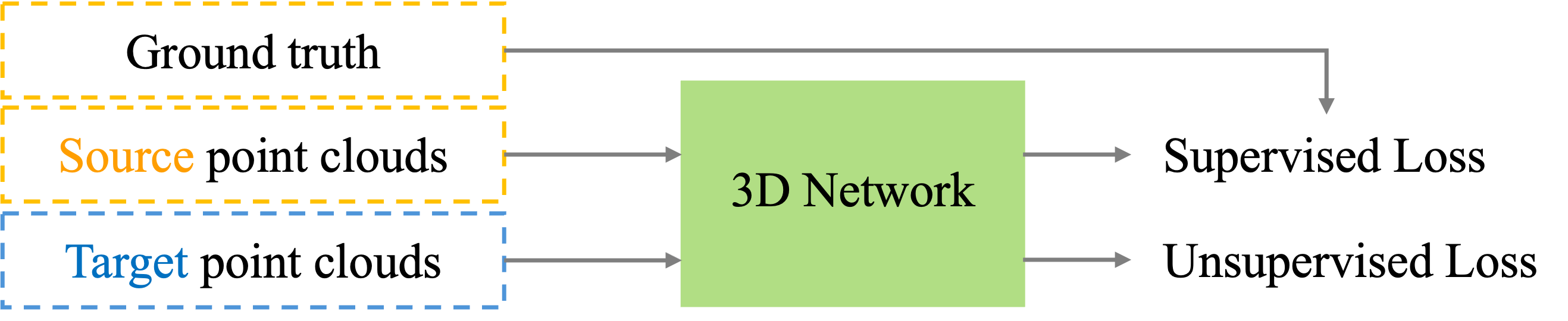}
    \vspace{-10pt}
    \caption{Typical UDA pipeline for 3D network training}
    \label{fig:uda}
    \vspace{-15pt}
\end{figure}

\vspace{-5pt}
\subsubsection{Problem setup}\label{subsec.uda_set}
Given source-domain point clouds $X^S$ with the corresponding labels $Y^S$ and target-domain point clouds $X^T$ without labels, the goal of point cloud adaptation is to learn a model $F$ that can produce accurate predictions $\hat{Y}^T$ for unseen target data. The network training in UDA consists of two typical learning tasks, i.e., supervised learning from the labelled source data and unsupervised adaptation toward unlabelled target data, as shown in Fig.~\ref{fig:uda}. Adaptation is usually achieved via four learning approaches: adversarial training, self-training, self-supervised learning, and style transfer.

\textit{Adversarial training}~\cite{tzeng2017adversarial} aims to learn domain-invariant features by training the model to extract features (from both source and target samples) that are indistinguishable by a domain discriminator; \textit{Self-Training}~\cite{zou2018unsupervised} employs a source-trained model to pseudo-label target data and uses confident target predictions to iteratively retrain the model, assuming that confident target predictions are correctly labelled; \textit{Self-Supervised Learning}~\cite{xiao2022unsupervised} learns useful representations from unlabelled target data by defining tasks solvable without human annotations, helping the network tolerate domain shifts and improve generalization; \textit{Style Transfer}~\cite{hoffman2018cycada} aims to translate source data to resemble target data for model training, which learns a mapping function that transforms the source data to have similar styles as the target data. The following subsections review domain adaptive point cloud learning for various 3D tasks. Table~\ref{tab. Sum of domain adaptation methods} provides an overview of representative methods.

\vspace{-5pt}
\subsubsection{Domain adaptation for 3D shape classification}\label{DAdapt-cls}

\begin{figure}[t]
    \centering
    \includegraphics[width=\linewidth]{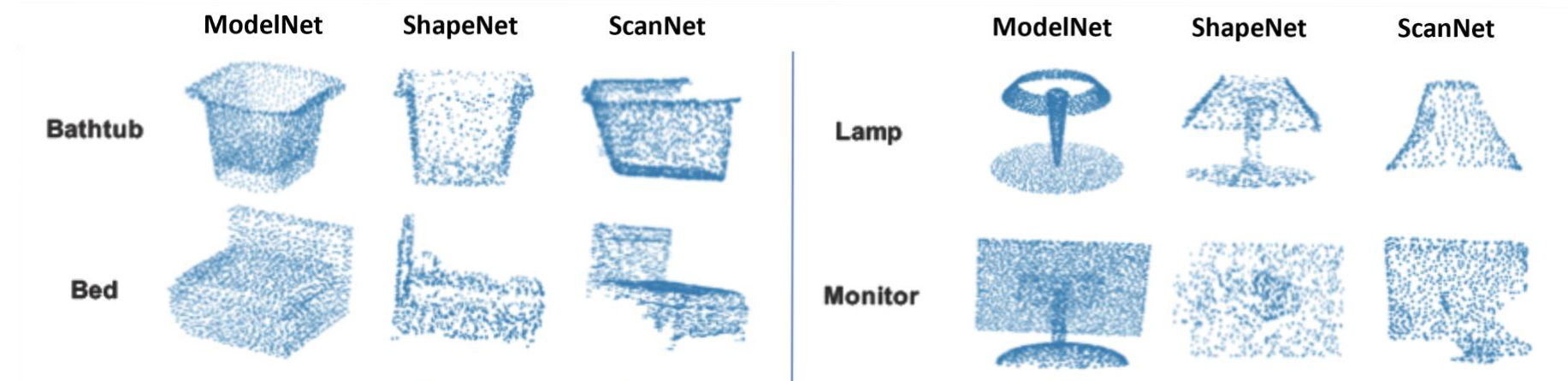}
    \vspace{-10pt}
    \caption{Examples of object-level point clouds in datasets ModelNet~\cite{wu20153d}, ShapeNet~\cite{chang2015shapenet}, and ScanNet~\cite{dai2017scannet}. The figure is reproduced based on \cite{qin2019pointdan}.}
    \label{fig:pointdan}
    \vspace{-5pt}
\end{figure}

Object-level point clouds are often collected from various sources, such as synthetic CAD models~\cite{wu20153d,chang2015shapenet} and real 3D scans~\cite{uy2019revisiting,dai2017scannet}, resulting in geometric discrepancies as shown in Fig.~\ref{fig:pointdan}. Recent studies have explored UDA for 3D shape classification across different 3D object datasets.

PointDAN~\cite{qin2019pointdan} uses adversarial training and Maximum Classifier Discrepancy\cite{saito2018maximum} to align features across domains. Subsequent studies~\cite{zou2021geometry,shen2022domain,liang2022point} focused on self-paced self-training for domain adaptation, gradually lowering the confidence threshold for selecting pseudo labels. Fan et al.\cite{fan2022self} developed a voting strategy to pseudo-label target samples by finding the nearest source neighbors in a shared feature space. Chen et al.\cite{chen2022quasi} proposed quasi-balanced self-training to address class imbalance in pseudo-labelling. Cardace et al.~\cite{cardace2021refrec} refined noisy pseudo-labels by matching shape descriptors learned through unsupervised shape reconstruction tasks in both domains.

SSL tasks were also designed to help networks learn domain-invariant features from unlabelled point cloud objects. Zou et al.\cite{zou2021geometry} introduced a joint task to predict rotation angles and distortion locations. Fan et al.\cite{fan2022self} reconstructed the squeezed 2D projections of objects back to 3D space. Shen et al.~\cite{shen2022domain} learned unsupervised features by approximating unsigned distance fields.

\vspace{-5pt}
\subsubsection{Domain adaptation for 3D object detection}\label{DAdapt-det}

Scene-level point clouds face significant geometric shifts due to variations in physical environments, sensor configurations, and weather conditions, making domain adaptation for 3D object detection particularly challenging. This area has garnered increased attention due to its importance.

The pioneering work by Wang et al.~\cite{wang2020train} showcased the efficacy of car size normalization in enhancing 3D object detection across countries. Subsequently, adversarial training was investigated, with Su et al.~\cite{su2020adapting} improved adversarial training by disentangling domain-specific attributes from LiDAR semantic features. Zhang et al.~\cite{zhang2021srdan} further developed scale-aware and range-aware domain alignment strategies to enhance adversarial training by leveraging the geometric properties of LiDAR point clouds.

Several methods have explored self-training for domain adaptive 3D detection~\cite{caine2021pseudo,yang2021st3d,luo2021unsupervised,yang2022st3d++}. For instance, ST3D~\cite{yang2021st3d} enhances pseudo labels using a quality-aware triplet memory bank and trains networks with curriculum data augmentation. Luo et al.~\cite{luo2021unsupervised} developed a multi-level consistency network ensuring consistency across points, instances, and neural statistics. Some studies~\cite{saleh2019domain,hahner2021fog,xu2021spg,hahner2022lidar} explored style transfer, such as simulating weather conditions over authentic point clouds to mitigate domain discrepancy~\cite{hahner2021fog, hahner2022lidar}. Xu et al.~\cite{xu2021spg} generated semantic points in missing object parts to enhance detection across domains. Yihan et al.~\cite{yihan2021learning} proposed a 3D contrastive co-training approach, while Wei et al.~\cite{wei2022lidar} introduced a teacher-student framework to bridge the domain gap caused by different LiDAR beam configurations.

\vspace{-5pt}
\subsubsection{Domain adaptation for 3D segmentation}\label{DAdapt-seg}
\begin{figure}
    \centering
    \includegraphics[width=\linewidth]{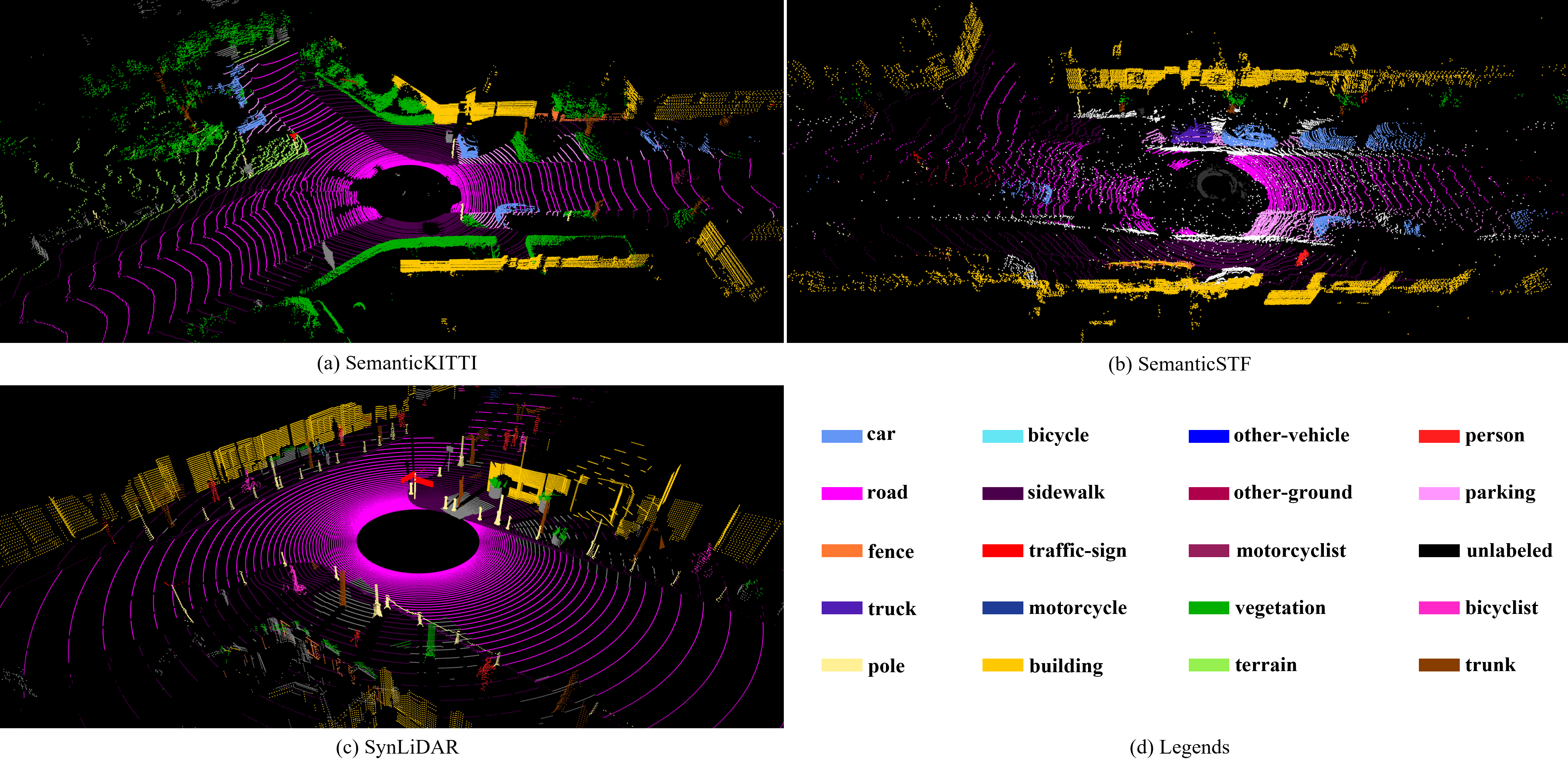}
    \vspace{-5pt}
    \caption{Example of LiDAR scans of different domains.  (a) A real scan of normal weather in SemanticKITTI~\cite{behley2019semantickitti}, (b) A real scan of adverse weather of snow in SemanticSTF~\cite{xiao20233d}, and (c) A synthetic scan in SynLiDAR~\cite{xiao2022transfer}. Different colors denote different semantic categories as in (d).}
    \label{fig:DA_synlidar2kitti}
    \vspace{-10pt}
\end{figure}

LiDAR point clouds often have significant domain discrepancies due to variations in physical environments, sensor configurations, weather conditions, etc. Hence, most prior UDA studies~\cite{yi2021complete,xiao2022transfer,saltori2022cosmix,saltori2022gipso,Liu_2023_CVPR,ryu2023instant,xiao2023domain,xiao2024domain} focus on outdoor LiDAR point clouds, while just a few~\cite{ding2022doda} tackle the issue for indoor point clouds. Fig.~\ref{fig:DA_synlidar2kitti} shows point-cloud samples of different domains that have clear domain discrepancies.

Studies on domain adaptive point cloud semantic segmentation can be broadly classified into two categories namely, uni-modal UDA that works with point clouds alone and cross-modal UDA that employs both point clouds and image data in training.

For \textit{uni-modal UDA}, a line of studies~\cite{wu2019squeezesegv2,langer2020domain, zhao2021epointda,jiang2021lidarnet,Li_2023_CVPR} projected point clouds to depth images and adopted 2D UDA methods to mitigate domain shifts. For example, Li et al.~\cite{Li_2023_CVPR} proposed an adversarial training framework to learn to generate source image masks to mimic the pattern of irregular target noise.
However, the 3D-to-2D projection loses geometric information, and most 2D UDA methods cannot handle the unique geometry of point clouds. Moreover, most 2D UDA methods adopt CNN architectures and cannot be generalized to point cloud architectures. 

Another line of UDA methods performed directly over point clouds. Yi et al.\cite{yi2021complete} transformed domain adaptation into a 3D surface completion task. Xiao et al.\cite{xiao2022transfer} used GANs to translate synthetic point clouds to match the sparsity and appearance of real ones. Saltori et al.\cite{saltori2022cosmix} and Xiao et al.\cite{xiao2022polarmix} mixed point clouds from source and target domains to create intermediate representations with reduced domain discrepancy.

\begin{table*}[h]
    \scriptsize
    \setlength\tabcolsep{1pt}
    \centering
    \caption{Summary of domain generalization methods.}
    \vspace{-5pt}
    \begin{tabular}{|l|c|c|l|}
    \hline
        Method & Published in & Task & Contribution \\
    \hline
        MetaSets~\cite{huang2021metasets} & CVPR 2021 & Classification & Meta-learning for generalized 3D features by classification on transformed point sets with geometric priors. \\
        MAL~\cite{huang2022manifold} & ECCV 2022 & Classification & Manifold adversarial learning for domain-generalized 3D representations. \\
    \hline
        3D-VField~\cite{lehner20223d} & CVPR 2021 & Detection & Adversarial augmentation for generalized 3D object detection over crashed vehicles.\\
        DG-BEV~\cite{wang2023towards} & CVPR 2023 & Detection & A domain generalized multi-view 3D object detection in BEV. \\
        Eskandar et al.\cite{eskandar2024empirical} & CVPR 2024 & Detection & An empirical study on architecture, voxel encoding, data augmentations, and anchor strategies. \\
    \hline
        PointDR~\cite{xiao20233d} & CVPR 2023 & Semantic Segmentation & Contrastive learning for domain generalizable 3D segmentation across different weather conditions.\\
        DGLSS~\cite{Kim_2023_CVPR} & CVPR 2023 & Semantic Segmentation & Learn density-variation representations that are tolerant to 3D sensor changes. \\
        3DLabelProp~\cite{sanchez2023domain} & ICCV 2023 & Semantic Segmentation & Domain generalized segmentation for LiDAR sequential point clouds. \\
        BEV-DG~\cite{li2023bev} & ICCV 2023 & Semantic Segmentation & BEV-based area-to-area fusion for 2D-3D cross-modal learning of segmentation.\\
        UniMix~\cite{zhao2024unimix} & CVPR 2024 & Semantic Segmentation & Mixing across weather over spatial, intensity, and semantic distributions. \\
    \hline
    \end{tabular}
    \label{tab. Sum of domain generalization methods}
\end{table*}

For \textit{cross-modal UDA}, each training sample typically comprises a 2D image and a 3D point cloud that are synchronized across LiDAR and camera sensors. Point-wise 3D annotations are provided for source data. The goal is to learn a robust point cloud segmentor that can work independently and requires no images for testing. Though the paired images can enrich the learned representation, cross-modal UDA faces new challenges due to the heterogeneity of the input spaces for images and point clouds as well as additional domain shifts between source and target images. Jaritz et al.~\cite{jaritz2020xmuda} developed xMUDA, the first cross-modal UDA framework that adopts a two-stream architecture to address the domain gap of each modality individually. Peng et al.~\cite{peng2021sparse} achieved cross-modal UDA with two modules, the first employing intra-domain cross-modal learning for cross-modal interaction while the second adopting adversarial learning for cross-domain feature alignment via inter-domain cross-modal learning.

Further, the exploration of domain adaptive \textit{3D instance segmentation} is largely neglected. Though Be{\v{s}}i{'c} et al.~\cite{bevsic2022unsupervised} explored domain adaptative 3D panoptic segmentation, the specific task of point-cloud instance segmentation remains unexplored. More research endeavors are imperative on instance-level point cloud segmentation for better understanding and exploitation of point cloud data.

\vspace{-5pt}
\subsubsection{Extension}\label{subsec.uda_extension}

\noindent \textbf{Source-free UDA}~\cite{huang2021model} is a variant of UDA that aims to adapt source-trained models to target distributions without accessing the source data in training. It is useful when data privacy and data portability are critical. 
Saltori et al.~\cite{saltori2022gipso} introduced an adaptive self-training method with geometric-feature propagation for source-free UDA of 3D LiDAR segmentation.

\noindent \textbf{Test-time domain adaptation} (TTA) is a setup where a source-pretrained model is adapted using only the unlabelled test data, usually with a \textit{single epoch} of training. Unlike typical UDA, the goal of TTA is to avoid collecting target data in advance, where the model is adapted with the test data flow. Though TTA is practical in real-world scenarios, it is challenging as the target data is available in test-stage only. 
Recently, Inkyu Shin et al.~\cite{shin2022mm} introduced a TTA method for multi-modal 3D semantic segmentation with a modal fusion module for more accurate segmentation.

\subsection{Domain generalization}\label{subsec.domain_generalization}
Unlike UDA, domain generalization (DG)~\cite{zhou2022domain} eliminates the dependency on target training data, making it highly valuable for real-world tasks where obtaining target data is difficult or costly prior to model deployment. This is crucial for many point cloud tasks that require 3D deep models to be robust and generalizable to unseen domains, such as in safe autonomous driving. Table~\ref{tab. Sum of domain generalization methods} provides an overview of representative methods.

\begin{figure}[t]
    \centering
    \includegraphics[width=0.9\linewidth]{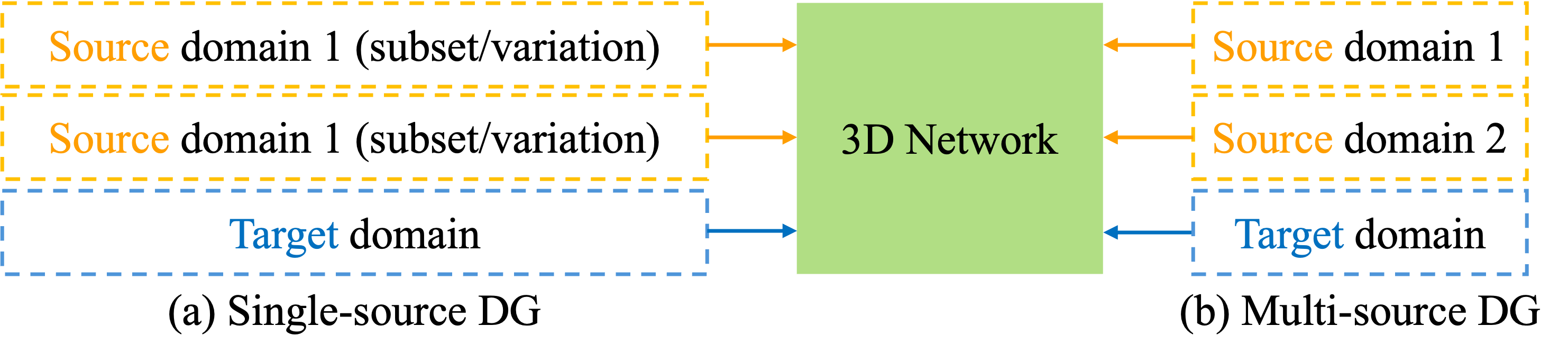}
    \vspace{-10pt}
    \caption{Typical pipeline of domain generalization (DG) including (a) Single-source DG; (b) Multi-source DG.}
    \label{fig:dg}
    \vspace{-10pt}
\end{figure}

\vspace{-5pt}
\subsubsection{Problem setup}

Given labelled point clouds of $K$ similar but distinct source domains $\mathcal{S}={\{S_k=\{(x^{(k)},y^{(k)})\}\}}_{K}^{k=1}$, where $x$ denotes a point cloud and $y$ is its labels, DG aims to learn a deep model $F$ with the source data only that can perform well in unseen target domain $\mathcal{T}$. Similar to 2D DG studies~\cite{zhou2022domain}, we review two DG settings for 3D point clouds as shown in Fig.~\ref{fig:dg}.
The first is multi-source DG which assumes the availability of more than one source domain in training, i.e., $K>1$. The motivation is to learn domain-invariant features (from multiple similar but distinct source domains) that can generalize well to any unseen domains. The second is single-source DG which is more challenging as it allows training data from a single source domain only. At the other end, single-source DG methods are more generic and can be applied to multi-source DG problems by ignoring the domain label.

\vspace{-5pt}
\subsubsection{Domain generalization for 3D shape classification}\label{DG-cls}

Huang et al.\cite{huang2021metasets} introduced a meta-learning framework for handling geometry shifts from CAD (e.g., ModelNet\cite{wu20153d}) to real object point clouds (e.g., ScanObjectNN~\cite{uy2019revisiting}). Later, Huang et al.~\cite{huang2022manifold} proposed manifold adversarial training using geometric transformations to generate intermediate domain samples. Both studies fall under the single-source DG setting.

\begin{table*}[h]
    \scriptsize
    \setlength\tabcolsep{1pt}
    \centering
    \caption{Summary of weakly-supervised methods. $^{\dag}$ denotes incomplete supervision; $^{\ddag}$ denotes inexact supervision; $^{\natural}$~denotes inaccurate supervision. ``D”, ``SS”, and ``IS” denote object detection, semantic segmentation, and instance segmentation, respectively.}
    \vspace{-5pt}
    \begin{tabular}{|l|c|c|l|l|}
    \hline
        Method & Publication & Task & Annotations & Contribution \\
    \hline
        BPCF\cite{tang2019transferable}& ICCV2019 & D & 2D boxes$^{\ddag}$ & Train a 3D objector using 2D\&3D boxes for strong classes, and only 2D boxes for weak classes.\\
        VS3D\cite{qin2020weakly} & {\tiny ACM-MM2020} & D & 2D images$^{\ddag}$ & Unsupervised 3D proposal and 2D-3D cross-modal knowledge distillation for final prediction.\\
        WS3D\cite{meng2020weakly} & ECCV2020 & D & Position-level & Proposing cylindrical objects from massive horizontal object centers click-annotated in BEV, \vspace{-2pt}\\
        & & & \&instances$^{\dag\ddag\natural}$ & and refining the detector with a few precise instance labels.\\
        WyPR\cite{ren20213d} & CVPR2021 & D/SS & Scene-level$^{\ddag}$ & Jointly learn point-wise segmentation and detection of 3D boxes by using scene-level class tags. \\
        BR\cite{xu2022back} & CVPR2022 & D & Position-level$^{\ddag}$ & Using synthetic 3D shapes to convert weak object center labels into fully-annotated virtual scenes for training.\\
        SS3D\cite{liu2022ss3d} & CVPR2022 & D & Sparse boxes$^{\dag}$ & Annotate one instance per scene with mining modules for sparse label learning.\\
        ViT-WS.\cite{zhang2023simple} & ICCV2023 & D & Point-level$^{\ddag}$ & A plain vision transformer trained on limited fully labelled and extensive weakly single-point labelled 3D objects.\\
        CoIn\cite{xia2023coin}  & ICCV2023 & D & Sparse boxes$^{\dag}$ & Contrastive instance feature mining for training detectors with limited annotations.\\
        NoiseDet\cite{chen2023learning}  & ICCV2023 & D & Noisy boxes$^{\natural}$ & Noise-resistant instance supervision for enhanced pseudo labelling with better generalization. \\
        Wang et al.\cite{wang2023not} & ICCV 2023 & D & Sparse boxes$^{\dag}$ & Side-aware framework with explicit side-specific localization quality estimation and importance assignment. \\
        DQS3D\cite{gao2023dqs3d} & ICCV 2023 & D & Sparse boxes$^{\dag}$ & A single stage densely-matched quantization-aware semi-supervised 3D Detection framework.\\
    \hline
        WSPCS\cite{xu2020weakly} & CVPR2020 & SS & Sparse points$^{\dag}$ & Train segmentation model with a small fraction (10\%) of labelled points.\\
        MPRM\cite{wei2020multi} & CVPR2020 & SS & Cloud-level$^{\dag}$ & Multi-path region mining to generate dense point-level labels from a classification network. \\
        1T1C\cite{liu2021one} & CVPR2021 & SS & Sparse points$^{\dag}$ & Label one point per object and self-train segmentation model with sparse labels. \\
        CSC\cite{hou2021exploring} & CVPR2021 & SS & Sparse points$^{\dag}$ & Unsupervised contrastive pre-training followed by fine-tuning with $<0.1\%$ labelled points.\\
        WSSS\cite{zhang2021weakly} & AAAI2021 & SS & Sparse points$^{\dag}$ & Self-supervised pre-training with point cloud colorization and sparse label propagation for class prototype learning. \\
        PNAL\cite{ye2021learning} & ICCV2021 & SS & Noisy points$^{\natural}$ & Noise-rate blind framework with point confidence selection and cluster label correction.\\
        PSD\cite{zhang2021perturbed} & ICCV2021 & SS & Sparse points$^{\dag}$ & Self-distillation for perturbation consistency and context-aware learning for affinity context.\\
        Redal\cite{wu2021redal} & ICCV2021 & SS & Sparse points$^{\dag}$ & Active learning with diversity-aware label selection for region segmentation. \\
        Scribble\cite{unal2022scribble} & CVPR2022 & SS & Scribbles$^{\ddag}$ & Scribble-supervised LiDAR segmentation. \\
        Hyb.CR\cite{li2022hybridcr} & CVPR2022 & SS & Sparse points$^{\dag}$ & Weakly-supervised 3D segmentation with point consistency and contrastive learning. \\
        TMSGP\cite{shi2022weakly} & CVPR2022 & SS & Sparse points$^{\dag}$ & Annotate 0.1\% points in the first frame of LiDAR sequences for temporal-spatial learning. \\
        MIL\cite{yang2022mil} & CVPR2022 & SS & Sparse points$^{\dag}$ & A 3D Transformer with inter-cloud learning and adaptive global weighted pooling. \\
        SQN\cite{hu2022sqn} & ECCV2022 & SS & Sparse points$^{\dag}$ & A point neighborhood query network leveraging 1\% sparse point annotations for training.\\
        DAT\cite{wu2022dual} & ECCV2022 & SS & Sparse points$^{\dag}$ & Adversarial dual adaptive transformations enforcing local and structural smoothness constraints. \\
        WS3D\cite{liu2022weakly} & ECCV2022 & SS & Sparse points$^{\dag}$ & Energy-based loss with boundary awareness and unsupervised region-level semantic contrast. \\
        LESS\cite{liu2022less} & ECCV2022 & SS & Sparse points$^{\dag}$ & Heuristic pre-segmentation for annotating, prototype learning, and multi-scan distillation. \\
        Lidal\cite{hu2022lidal} & ECCV2022 & SS & Sparse points$^{\dag}$ & Active learning by using inter-frame divergence and entropy as the selection metrics. \\
        CPCM\cite{liu2023cpcm} & CVPR2023 & SS & Sparse points$^{\dag}$ & Contextual 3D modeling with a region-wise masking and a contextual masked training. \\
        HPAL\cite{xu2023hierarchical} & ICCV2023 & SS & Sparse points$^{\dag}$ & A hierarchical point-based active learning method for semi-supervised point cloud semantic segmentation. \\
    \hline
        SPIB\cite{liao2021point} & {\tiny T-PAMI2021} & IS & Boxes\&points$^{\dag\ddag}$ & Semi-supervised 3D proposal generation and semantic propagation for instance prediction.\\
        B2M\cite{chibane2022box2mask} & ECCV2022 & IS & Box-level$^{\ddag}$ & 3D bounding box voting and instance clustering. \\
        GaPro\cite{ngo2023gapro} & ICCV2023 & IS & Box-level$^{\ddag}$ & Pseudo labelling from box annotations and self-training for instance segmentation.\\
        FreePoint\cite{zhang2023freepoint} & CVPR2024 & IS & Sparse points$^{\dag}$ & Clustering for unsupervised class-agnostic IS and weakly-supervised pre-training IS.\\
    \hline
    \end{tabular}
    \label{tab. Sum of weakly-supervised methods}
\end{table*}

\vspace{-5pt}
\subsubsection{Domain generalization for 3D object detection}\label{DG-det}

Improving 3D detector generalizability for unseen domains is crucial for tasks like autonomous driving, but DG for 3D object detection is relatively under-explored. Lehner et al.~\cite{lehner20223d} made the first attempt at single-source DG with an adversarial augmentation method to deform point clouds during training. Recently, Wang et al.~\cite{wang2023towards} proposed a single-source DG approach for multi-view 3D object detection in Bird-Eye-View (BEV). It decouples depth estimation from camera parameters, using dynamic perspective augmentation, and adopting multiple pseudo-domains for better generalization.

\vspace{-5pt}
\subsubsection{Domain generalization for 3D segmentation}\label{DG-seg}

Several studies on domain-generalizable point cloud semantic segmentation have been reported recently. Xiao et al.\cite{xiao20233d} study outdoor point cloud segmentation under adverse weather, where a domain randomization and aggregation learning pipeline was designed to enhance the model generalization performance. \cite{Kim_2023_CVPR} augments the source domain and introduces constraints in sparsity invariance consistency and semantic correlation consistency for learning more generalized 3D LiDAR representations.

Similar to the domain adaptation, domain-generalizable 3D instance segmentation remains under-explored.

\subsection{Summary}
Transferring knowledge across domains is crucial for maximizing the use of existing annotations, leading to extensive studies of UDA and DG in machine learning. However, despite significant advancements in areas like 2D vision and NLP, UDA and DG for point clouds remain under-explored, as indicated by fewer publications and lower benchmark performance. Consequently, more efforts are urgently needed to advance this promising research area.

\section{Weakly-supervised learning}\label{sec.weakly}

Weakly-supervised learning (WSL), as an alternative to fully-supervised learning, leverages weak supervision for network training. 
Collecting weak annotations often reduces the annotation cost and time significantly, making WSL an important branch of label-efficient learning.
With the WSL definition in~\cite{zhou2018brief}, we categorize WSL methods on point clouds based on three types of weak supervision: \textit{incomplete} supervision, \textit{inexact} supervision, and \textit{inaccurate} supervision. Incomplete supervision involves only a small portion of training samples being labelled, while inexact supervision provides coarse-grained labels that may not match the model output. Inaccurate supervision refers to noisy labels. Table~\ref{tab. Sum of weakly-supervised methods} shows a summary of representative approaches.

\subsection{Incomplete supervision}\label{subsec.incomplete_sup}
\begin{figure}[t]
    \centering
    \includegraphics[width=0.9\linewidth]{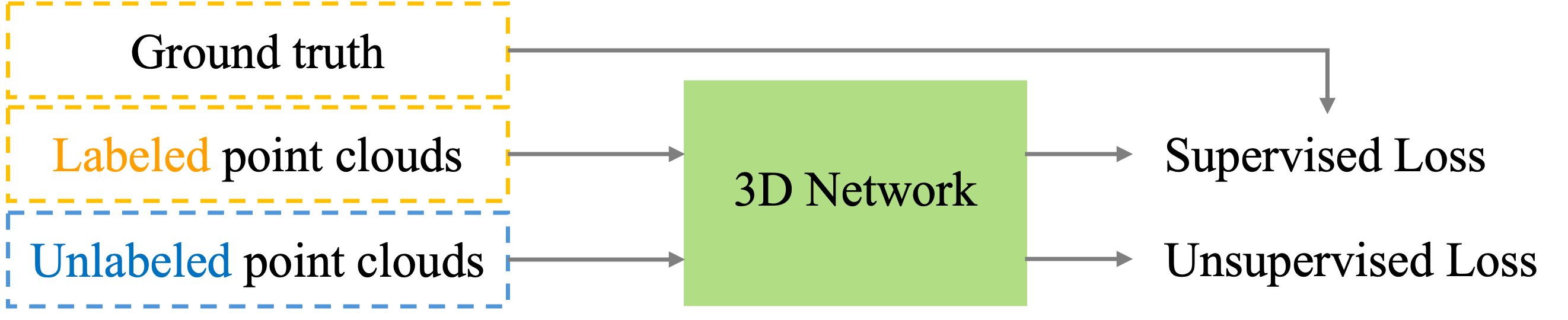}
    \vspace{-5pt}
    \caption{Typical pipeline of training 3D point cloud networks with incomplete supervision.}
    \label{fig:weakly-incomplete}
    \vspace{-4mm}
\end{figure}

In the context of incomplete supervision, only a subset of training point clouds is labelled. 
Incomplete supervision can be obtained with two labelling strategies: 1) sparsely labelling a small number of points from a large number of point-cloud frames and 2) intensively labelling a small number of point cloud frames with more (or fully) labelled points. Following conventions in relevant literature, we refer to studies with the first strategy by "3D weakly-supervised learning" and review them in Sec. \ref{subsec.weakly}. For the second strategy, we refer to it by "3D semi-supervised learning" and review relevant studies in Sec. \ref{subsec.semi}. Both labelling strategies adopt similar training paradigms with supervised learning on limited labelled points and unsupervised learning on massive unlabelled points, as shown in Fig.~\ref{fig:weakly-incomplete}. Additionally, we review "3D few-shot learning" in Sec. \ref{subsec.few-shot} with a few labelled samples of novel classes and many labelled samples of base classes, aiming to reduce the labelling of novel classes in network training.

\vspace{-5pt}
\subsubsection{3D weakly-supervised learning}\label{subsec.weakly}
3D weakly-supervised learning (3D-WSL) learns with a small number of sparsely annotated points in each point cloud frame. It has high research and application value since it allows annotating more point-clouds frames with less labelling redundancy.

\noindent\textbf{Problem setup}. Let $P$ be a point cloud of the training set consisting of labelled points $\{(X_l,Y_l)\}$ and unlabelled points $\{(X_u,\varnothing)\}$, where $X$ represents point space and $Y$ means label space. 3D-WSL aims to learn a function $f: X_l\cup X_u\mapsto Y$ given a large amount of point clouds including a tiny fraction of labelled points (e.g., 5\%) as training input.

\noindent \textbf{3D-WSL for semantic segmentation.} This task aims to develop robust segmentation models using only a small fraction of labelled points within each point cloud. One approach is through \textit{consistency-learning}~\cite{xu2020weakly,zhang2021perturbed,wu2022dual,wang2022new}, which enforces prediction consistency across different augmented views of the same input data. For example, Xu et al. \cite{xu2020weakly}  introduced a Siamese self-supervision branch for consistent learning from unlabelled points. Studies have explored \textit{contrastive learning}~\cite{liu2022weakly,li2022hybridcr} on unlabelled point clouds, pulling features of points towards their augmented views and away from other points to learn structural representations unsupervisedly. For example, Liu et al.\cite{liu2022weakly} segmented point clouds to extract boundaries for region-level contrastive learning. \textit{self-training}~\cite{liu2021one, shi2022weakly} has also been investigated. For example, Shi et al.\cite{shi2022weakly} annotated only a small portion of points in the first LiDAR frame and selected confident predictions of unlabelled points as pseudo labels for network re-training. Hu et al.\cite{hu2022sqn} recently proposed a Semantic Query Network that leverages sparsely labelled points and their local neighbours to learn a compact neighbourhood representation.

Instead of random selection, \textit{active learning} identifies more representative points for labelling. For example, Wu et al.\cite{wu2021redal} segmented point clouds based on entropy, color discontinuity, and structural complexity to select representative sub-regions. Hu et al.\cite{hu2022lidal} used prediction inconsistency across LiDAR frames to measure uncertainty for active sample selection. Additionally, recent studies~\cite{liu2022less, tao2022seggroup} actively labelled segments instead of individual points, reducing labelling efforts by pre-segmenting LiDAR sequences into connected components for coarse labelling, given the homogeneity of 3D objects in scenes.

\noindent \textbf{3D-WSL for object detection}. 3D-WSL for object detection is largely under-explored. Liu et al.~\cite{liu2022ss3d} recently conducted a pioneering exploration by annotating only one 3D object in each scene and then using the prediction confidence to mine object instances for network re-training.

\noindent \textbf{3D-WSL for instance segmentation}. 3D-WSL has recently been explored for instance segmentation by ``annotating one point per instance"~\cite{tao2022seggroup,tang2022learning}. For example, Tao et al.~\cite{tao2022seggroup} first over-segmented point clouds and then clicked one point per segment to assign its location, category, and instance identity.

\vspace{-5pt}
\subsubsection{3D semi-supervised learning}\label{subsec.semi}
3D semi-supervised learning (3D-SemiSL) works with intensive (full) annotation of a small portion of point cloud frames. As studied in~\cite{xu2020weakly}, the annotation strategy in 3D-SemiSL leads to inferior part segmentation than that in 3D-WSL, largely due to its higher annotation redundancy. However, 3D-SemiSL is advantageous in requiring less training data collection during annotations.

\noindent\textbf{Problem setup}. Given point clouds $\mathbf{X}_l\in\mathbb{R}^{N_l\times3}$ with labels $\mathbf{Y}_l$ and unlabelled point clouds $\mathbf{X}_u\in\mathbb{R}^{N_u\times3}$ ($N_l$ and $N_u$ are point cloud numbers, $N_l<N_u$), 3D-SemiSL aims to learn a point cloud model $F$ from the labelled data and unlabelled data that can perform well on unseen point clouds.

\noindent\textbf{3D-SemiSL for object detection}. 
Most existing studies adopted the Mean-Teacher framework~\cite{tarvainen2017mean} consisting of a teacher network and a student network of the same architecture. The teacher model is a moving average of student models, and its predictions guide the student's learning. This setup assumes the teacher model learns more robust representations, which benefit the student model.

SESS~\cite{zhao2020sess} employs consistency learning~\cite{verma2019interpolation} between the teacher and student networks, aiming for a perturbation-invariant output distribution by assuming that decision boundaries lie in low-density regions. 3DIoUMatch~\cite{wang20213dioumatch} uses confident predictions from the teacher model as pseudo labels for re-training, minimizing the entropy of student model predictions~\cite{grandvalet2004semi} and lowering point density at decision boundaries~\cite{lee2013pseudo}. Yin et al.~\cite{yin2022semi} proposed the Proficient Teacher model, introducing a spatial-temporal ensemble module and a clustering-based box voting strategy to enhance pseudo-labelling of 3D bounding boxes. Liu et al.~\cite{Liu_2023_CVPR} introduced a dual-threshold strategy and data augmentation to improve hierarchical supervision and feature representation in training the student network.

\noindent\textbf{3D-SemiSL for segmentation}. 
Compared to 3D bounding box annotations, point-wise labelling is even more laborious and time-consuming. Consequently, 3D-SemiSL for point cloud segmentation has garnered significant attention recently~\cite{jiang2021guided,cheng2021sspc,deng2022superpoint,kong2022lasermix,chu2022twist}.

For \textit{3D semantic segmentation}, Jiang et al.\cite{jiang2021guided} proposed a guided point contrastive learning framework to improve model generalization. Cheng et al.\cite{cheng2021sspc} developed superpoint graphs and used pseudo-labels for superpoints to train graph neural networks semi-supervisedly. Kong et al.\cite{kong2022lasermix} mixed laser beams from different LiDAR scans to encourage the model to make consistent predictions pre- and post-mixing, enhancing generalization. Li et al.\cite{li2023less} introduced Sparse Depthwise Separable Convolution, creating a lightweight segmentation model that requires less training data.
For \textit{3D instance segmentation}, Chu et al.~\cite{chu2022twist} proposed a two-way inter-label self-training framework that leverages pseudo semantic labels and pseudo instance proposals to mutually denoise pseudo signals for better semantic-level and instance-level supervision.

\noindent \textbf{3D-SemiSL for other tasks}. 3D-SemiSL has also been explored in other 3D point cloud tasks due to its efficiency in reducing human annotations, with applications in \textit{point cloud registration}\cite{huang2020feature}, \textit{hand pose estimation}\cite{chen2019so}, and more.

\noindent \textbf{Semi-supervised domain adaptation.} The combination of semi-supervised learning and domain adaptation, known as semi-supervised domain adaptation (SSDA), leverages labelled source samples, many unlabelled target samples, and a small amount of labelled target samples to train a model that performs well in the target domain. SSDA has been explored for various point cloud learning tasks, including semantic segmentation~\cite{xiao2022transfer} and 3D object detection~\cite{wang2023ssda3d}.

\vspace{-5pt}
\subsubsection{3D few-shot learning}\label{subsec.few-shot}

Fully supervised models learn from a large amount of training samples under a ``closed set" setup, i.e., the training and testing data have the same label space. Such supervised learning is not ideal for quickly learning new concepts with limited data, which motivates few-shot learning (FSL) that aims to learn a novel class from just a few labelled samples. FSL can be seen as an extension of semi-supervised learning in an "open-set" setup that has only a few labelled samples of novel classes, along with many labelled samples of base classes \cite{qi2020small}. Due to its superb merit in data requirements, FSL has recently attracted increasing attention.

\noindent\textbf{Problem setup.} There are two typical settings in FSL: The \textit{$N$-way-$K$-shot}~\cite{chencloser} where the training set and testing set are disjoint in terms of classes; The \textit{generalized FSL}~\cite{kang2019few} that recognizes both base and novel classes in testing. 
\begin{itemize}[leftmargin=*,noitemsep,topsep=0pt]
    \item \textbf{$N$-way-$K$-shot.} Let $(x, y)$ denotes a point cloud $x$ and its label $y$. FSL aims to train a model on a group of few-shot tasks sampled from a data set with a training class set $C_{\mathrm{train}}$ and test the trained model on another group of tasks sampled from a data set with new classes $C_{\mathrm{test}}$, where $C_{\mathrm{train}} \cap C_{\mathrm{test}} = \varnothing$. Each few-shot task is denoted by an \textit{episode}, which is instantiated as a $N$-way-$K$-shot task with a few \textit{query} samples and \textit{support} samples: The query samples form a query set $\mathcal{Q}=\{(x_i^{\mathcal{Q}},y_i^{\mathcal{Q}}))\}^{N_q=N\times Q}_{i=1}$ containing $N$ classes in $C_{\mathrm{train}}$ with $Q$ samples of each class, and the support samples forms a support set $\mathcal{S}=\{(x_i^{\mathcal{S}},y_i^{\mathcal{S}}))\}^{N_s=N\times K}_{i=1}$ containing the same $N$ classes in $C_{\mathrm{train}}$ with $K$ examples of each class. The goal of the $N$-way-$K$-shot learning is to train a model $F(x^{\mathcal{Q}},{\mathcal{S}})$ that predicts the label distribution $H$ for any query point cloud $x^{\mathcal{Q}}$ based on ${\mathcal{S}}$. In testing, the trained model is tested over the testing episodes ${\mathcal{V}} = {(S_j,Q_j)}^J_{j=1}$ for the new classes $C_{\mathrm{test}}$. Note the ground-truth labels $y^{\mathcal{Q}}$ of query samples are only available during training.
    \item \textbf{Generalized FSL.} This is a more challenging FSL setting. It involves training data of \textit{base} classes and \textit{novel} classes, including abundant labelled data of the base classes and few-shot labelled samples of the novel classes. The goal is to obtain a few-shot model that can learn to recognize novel objects by leveraging knowledge learnt from the base classes.
\end{itemize}

FSL has been widely studied for 2D images. Most work performs meta-learning in three typical approaches: 1) Metric learning~\cite{koch2015siamese} that measures the support-query similarity and group each query sample into its nearest support class in the latent space; 2) Optimization approach~\cite{ravi2017optimization} that differentiates support-set optimization for fast adaptation; 3) Model-based approach~\cite{santoro2016meta} that tailors model architectures for fast learning. We review 3D FSL for point clouds, a far under-explored area due to many modal-specific challenges such as unordered data structures, point sparsity, and large geometric variations. Several pioneering studies recently explored different 3D FSL tasks on 3D shape classification~\cite{ye2022makes,ye2022closer}, 3D semantic segmentation~\cite{zhao2021few,zhao2022crossmodal}, 3D object detection~\cite{zhao2022prototypical}, and 3D instance segmentation~\cite{ngo2022geodesic}.

\noindent \textbf{FSL for 3D shape classification.} Ye et al.~\cite{ye2022makes,ye2022closer} conducted a pioneering study on FSL for 3D shape classification under the \textit{$N$-way-$K$-shot} setting. They extended existing 2D FSL methods for 3D point-cloud data and introduced a baseline method to deal with high intra-class variance and subtle inter-class differences in point cloud representations. Yang et al. \cite{yang2023cross} projected point clouds into depth images and explored cross-modal FSL for 3D shape classification. In addition, Chowdhury et al. \cite{chowdhury2022few} studied \textit{few-shot class-incremental learning} that incrementally fine-tunes a trained model (on base classes) for novel classes with few samples.

\noindent\textbf{FSL for 3D Segmentation.}  Zhao et al. \cite{zhao2021few} explored FSL for 3D \textit{semantic segmentation} under the $N$-way-$K$-shot setting. They distill discriminative knowledge from scarce support that can effectively represent the distributions of novel classes and leverage such knowledge for 3D semantic segmentation.
An et al. \cite{an2024rethinking} propose a Correlation Optimization method for few-shot 3D point cloud segmentation.
Ngo et al. \cite{ngo2022geodesic} proposed a geodesic-guided transformer for 3D few-shot \textit{instance segmentation} of indoor dense point clouds. They employed a few support point cloud scenes and their ground-truth masks to generate discriminative features for instance mask prediction, and utilized geodesic distance as guidance with improved segmentation.

\noindent\textbf{FSL for 3D object detection.} FSL-based 3D detection is far under-explored. Zhao et al. \cite{zhao2022prototypical} designed Prototypical VoteNet, the first 3D detector for generalized FSL. They introduced a class-agnostic 3D primitive memory bank to store geometric prototypes of base classes and designed multi-head cross-attention to associate the geometric prototypes with scene points for better feature representations. 
Similar to 3D FSL segmentation, the study only covers indoor dense point clouds with dense representations.

\subsection{Inexact supervision}\label{subsec.inexact_sup}
The term ``inexact supervision" refers to supervision that is not as precise as desired for specific tasks. One example is coarse-grained labels that are much easier to collect.

\begin{figure}[t]
    \centering
    \includegraphics[width=\linewidth]{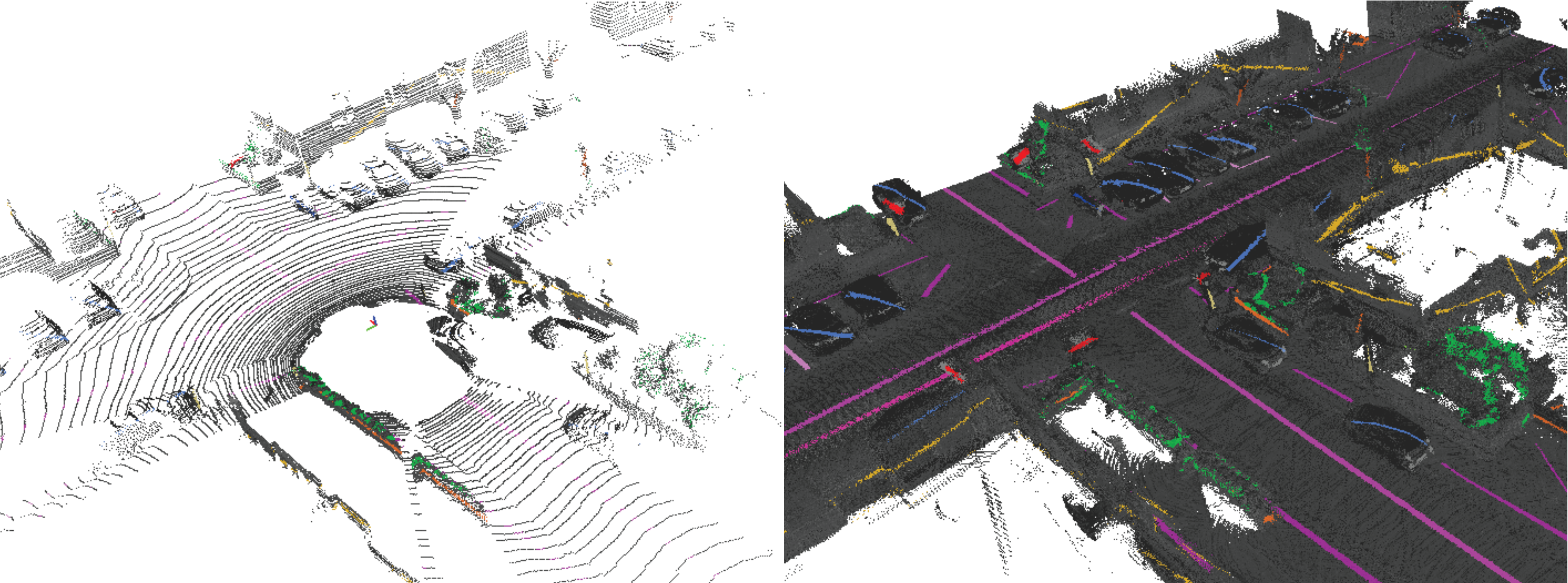}
    \vspace{-5pt}
    \caption{Example of scribble-annotated LiDAR point cloud scenes (left) and superimposed frames (right) in ScribbleKITTI~\cite{unal2022scribble}. The figure is extracted from \cite{unal2022scribble}.}
    \label{fig:ScribbleKITTI}
    \vspace{-5pt}
\end{figure}

\noindent \textbf{3D semantic segmentation.} Different weak supervision has been explored to save expensive point-wise annotation. For instance, Wei et al.\cite{wei2020multi,lin2022weakly} employed \textit{subcloud-level labels} for point cloud parsing, where classes appearing in the neighbourhood of uniformly sampled seeds are used as labels. Unal et al.\cite{unal2022scribble} used scribbles as labels for LiDAR point clouds as shown in Fig.~\ref{fig:ScribbleKITTI}, greatly facilitating the efficiency of point cloud labelling greatly.

\noindent \textbf{3D object detection.} Several recent studies used position-level annotations instead of 3D bounding boxes for 3D detection. For example, Meng et al.~\cite{meng2020weakly} and Xu et al.~\cite{xu2022back} used object centers to provide coarse position information for training. Ren et al.~\cite{ren20213d} employed scene-level tags for point cloud segmentation and detection without involving any point-wise semantic labels or object locations. Beyond 3D weak annotations, several studies exploited 2D image classes~\cite{liu2022eliminating} or 2D bounding boxes~\cite{wei2021fgr, qin2020weakly} to guide the training of 3D detectors. It reduces the annotation cost significantly as 2D annotations are much easier to collect.

\begin{figure}[t]
    \centering
    \includegraphics[width=0.9\linewidth]{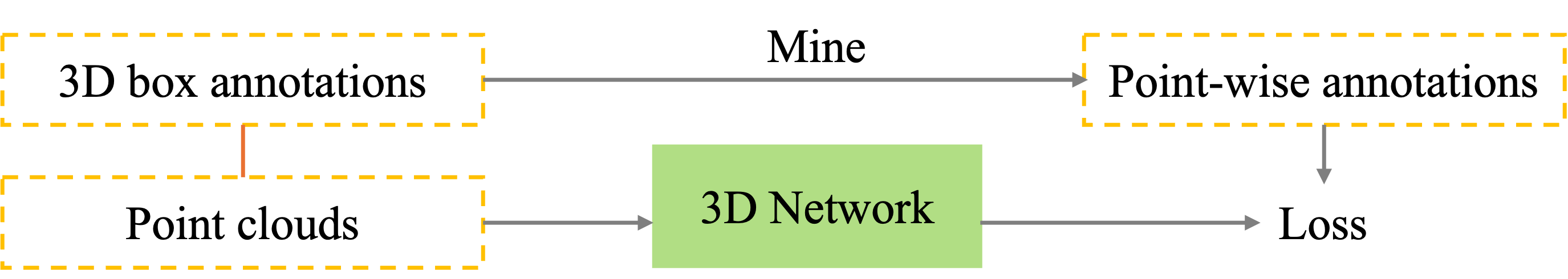}
    \vspace{-5pt}
    \caption{Typical pipeline of training 3D segmentation networks with inexact supervision of 3D bounding boxes.}
    \label{fig:weakly-inexact}
    \vspace{-10pt}
\end{figure}

\noindent \textbf{3D instance segmentation.} 
Another line of research~\cite{liao2021point,chibane2022box2mask} exploited coarse 3D bounding boxes to train 3D instance segmentation networks as illustrated in Fig.~\ref{fig:weakly-inexact}. To address the inaccuracy of 3D bounding boxes, Liao et al.~\cite{liao2021point} iteratively refined the bounding boxes and performed point-wise instance segmentation with the refined bounding boxes. Differently, Chibane et al.~\cite{chibane2022box2mask} introduced Box2Mask that adopts Hough Voting to generate accurate instance segmentation masks from 3D bounding boxes. These studies show promising 3D instance segmentation performance under weak supervision signals.

\subsection{Inaccurate supervision}\label{subsec.inaccurate_sup}
\begin{figure}[t]
    \centering
    \includegraphics[width=0.9\linewidth]{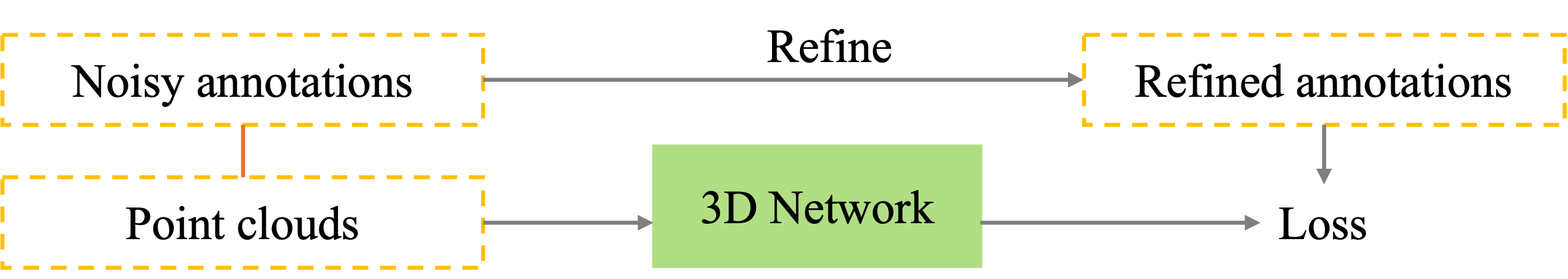}
    \vspace{-5pt}
    \caption{Typical pipeline for training 3D networks with inaccurate supervision}
    \label{fig:weakly-inaccurate}
    \vspace{-10pt}
\end{figure}

Inaccurate supervision refers to the annotations that are noisy with false labels. It’s very common as human annotators cannot guarantee 100\% accuracy especially when only limited time and resources are available. The noisy labels provide wrong guidance and often hinder network training. Therefore, the key to learning with inaccurate supervision is to refine annotations and improve supervision quality as illustrated in Fig.~\ref{fig:weakly-inaccurate}.

Despite the high research and application value, robust point cloud learning from inaccurate supervision is largely neglected in the literature. Ye et al.~\cite{ye2021learning,ye2022robust} designed a 3D semantic segmentation framework that introduces point-level confidence mechanism to select reliable labels and a cluster-level label correction process to refine training data. More research is needed to advance this very useful but far under-explored research area.

\subsection{Summary}
Weakly supervised point cloud learning aims to train robust deep models with limited, noisy, or imprecise annotations, addressing the challenge of preparing precise supervision for unordered and unstructured point clouds. This area has gained significant attention. Some studies show that weakly supervised models can achieve performance comparable to fully supervised ones. Researchers have also combined weakly supervised learning with techniques like transfer learning and self-supervised learning to enhance point cloud modeling. The ongoing progress in this field is expected to continue, with new methods regularly being proposed and evaluated, contributing to the broader advancement of 3D deep learning.

\section{Pretrained Foundation Models}\label{sec.pretraining}

\begin{table*}[ht]
    \vspace{-10pt}
    \scriptsize
    \setlength\tabcolsep{1pt}
    \centering
    \caption{Summary of methods based on multi-modal pre-trained foundation models.}
    \vspace{-5pt}
    \begin{tabular}{|l|c|l|}
    \hline
        Method & Publication & Contribution \\
    \hline
        PointCLIP\cite{zhang2022pointclip} & CVPR2022 & Zero-shot and few-shot 3D classification by aligning CLIP-encoded point cloud projections with category texts.\\
        PartSLIP\cite{liu2022partslip} & CVPR2023 & 
        Few-shot part segmentation by multi-view rendering and 2D detection through VLM (GLIP\cite{li2022grounded}) and text prompts.\\
        ULIP\cite{xue2022ulip}  & CVPR2023 & Pre-training for learning a unified representation of image, text, and 3D point cloud with object triplets contrastive learning. \\
        CLIP$^2$\cite{zeng2023clip2} & CVPR2023 & Contrastive language-image-point cloud pretraining by exploiting naturally-existed correspondences in 2D and 3D scenarios. \\
        OV-3DET\cite{lu2023open} & CVPR2023 & Open-vocabulary 3D object detection via text prompts by triplet cross-modal contrastive learning of point-clouds, images, and texts.\\
        P2W\cite{tang2023parts2words} & CVPR2023 & Learn joint embedding of point clouds and texts by bidirectional matching for shape-text retrieval.\\
        PLA\cite{ding2022language} & CVPR2023 & Caption images with VLMs for paired points and contrastive learn point-language representations for text-embedded 3D recognition.\\
        I2P-MAE\cite{zhang2023learning} & CVPR2023 &  Self-supervised pre-training for leveraging the pre-trained 2D models to guide 3D masked autoencoding.\\
        3D-VLP\cite{jin2023context}  & CVPR2023 & Point cloud-language pre-training via context-aware spatial-semantic alignment and mutual 3D-language masked modeling. \\
        CLIP2Scene\cite{chen2023clip2scene}  & CVPR2023 & Leverage CLIP for semantic-driven cross-modal contrastive pre-training for label-efficient 3D scene understanding.\\
        OpenScene\cite{peng2023openscene} & CVPR2023 & Learn features for 3D scene points that are co-embedded with texts and images in CLIP feature space for open-vocabulary queries.\\
        IDPT\cite{zha2023instance} & ICCV2023 & Instance-aware dynamic prompt tuning for pre-trained point cloud models for object-level 3D recognition.\\
        CLIP2Point\cite{huang2023clip2point} & ICCV2023 & Transfer CLIP to point cloud classification with image-depth pre-training. \\
        PT.CLIP V2\cite{zhu2023pointclipv2} & ICCV2023 & Collaborate CLIP and GPT to be a unified 3D open-world learner for zero-shot 3D classification, segmentation, and detection.\\
        UP-VL\cite{najibi2023unsupervised} & ICCV2023 & Auto-labelling amodal 3D bounding boxes and tracklets for open-set categories using 2D image-text pairs without 3D annotations.\\
        OpenMask3D\cite{takmaz2023openmask3d} & {\tiny NeurIPS2023} & Zero-shot open-vocabulary 3D instance segmentation by class-agnostic 3D instance mask prediction and aggregation of per-mask \vspace{-2pt}\\ 
        & & features via multi-view fusion of CLIP-based image embeddings. \\
        GeoZe\cite{mei2023geometrically} & CVPR2024 & Geometrically-driven aggregation approach for zero-shot point cloud understanding by leveraging VLMs.\\
        Open3DSG\cite{koch2024open3dsg} & CVPR2024 & Open vocabulary 3D scene graph learning from 3D point clouds with inter-object relationships extracted from grounded LLMs. \\
        PartDistill\cite{umam2023partdistill} & CVPR2024 & Knowledge distillation of 2D projected multi-view images from VLMs to facilitate 3D part segmentation. \\
        MaskClustering\cite{yan2024maskclustering} & CVPR2024 & A graph clustering based method to merge 2D mask  features from CLIP for open-vocabulary 3D instance segmentation (OV-3DIS). \\
        SAI3D\cite{yin2023sai3d} & CVPR2024 & Partitioning 3D scenes into geometric primitives, then merging into 3D segments consistent with multi-view SAM\cite{kirillov2023segment} masks. \\
        ULIP-2\cite{xue2023ulip2} & CVPR2024 & Tri-modal pre-training that leverages large multimodal models to automatically generate holistic language descriptions for 3D shapes. \\
        Open3DIS\cite{nguyen2023open3dis} & CVPR2024 & OV-3DIS aggregates 2D masks across frames into coherent point cloud regions and combines with 3D class-agnostic proposals. \\
        ZSVG3D\cite{yuan2023visual} & CVPR2024 & Zero-shot open-vocabulary 3D visual grounding that localizes 3D objects based on textual descriptions from large language models. \\
        LL3DA\cite{chen2023ll3da} & CVPR2024 & Large Language 3D assistant that takes point cloud as direct input and respond to both textual instructions and visual-prompts. \\
        TAMM\cite{zhang2024tamm} & CVPR2024 & A two-stage learning approach with three adapters: the CLIP Image Adapter bridges 3D-rendered and natural images, while Dual- \vspace{-2pt}\\ 
        & & Adapters separate 3D shape representation into visual attributes and semantic understanding for effective multi-modal pre-training. \\
    \hline
    \end{tabular}
    \label{tab. Sum of PFM methods}
\end{table*}

The recent advance of pretrained foundation models (PFMs) has yielded great breakthroughs across various AI fields including 2D computer vision, NLP, and their intersection (i.e., vision-language foundation models (VLMs)~\cite{zhou2023comprehensive,zhang2023vision}). PFMs with Internet-scale data in an unsupervised manner~\cite{he2020momentum,he2022masked, devlin2018bert} can be easily adapted to downstream tasks by fine-tuning with much fewer task data, enabling fast network convergence and learning with small data. In addition, VLMs~\cite{radford2021learning} trained with image-text pairs have demonstrated remarkable zero-shot visual prediction performance, being able to recognize objects of novel concepts with impressive accuracy without involving any labelled images but only text illustrations in training.

The great success of PFMs sheds light on label-efficient learning for point clouds. This section reviews related 3D studies with self-supervised pretraining in Sec. \ref{subsec.ssl} and multi-modal pretraining in Sec. \ref{subsec.multimodal}. Relevant challenges are finally discussed in Sec. \ref{subsec.pretrain-summary}. 

\subsection{Self-supervised pretraining}\label{subsec.ssl}

\begin{figure}[t]
    \centering
    \includegraphics[width=\linewidth]{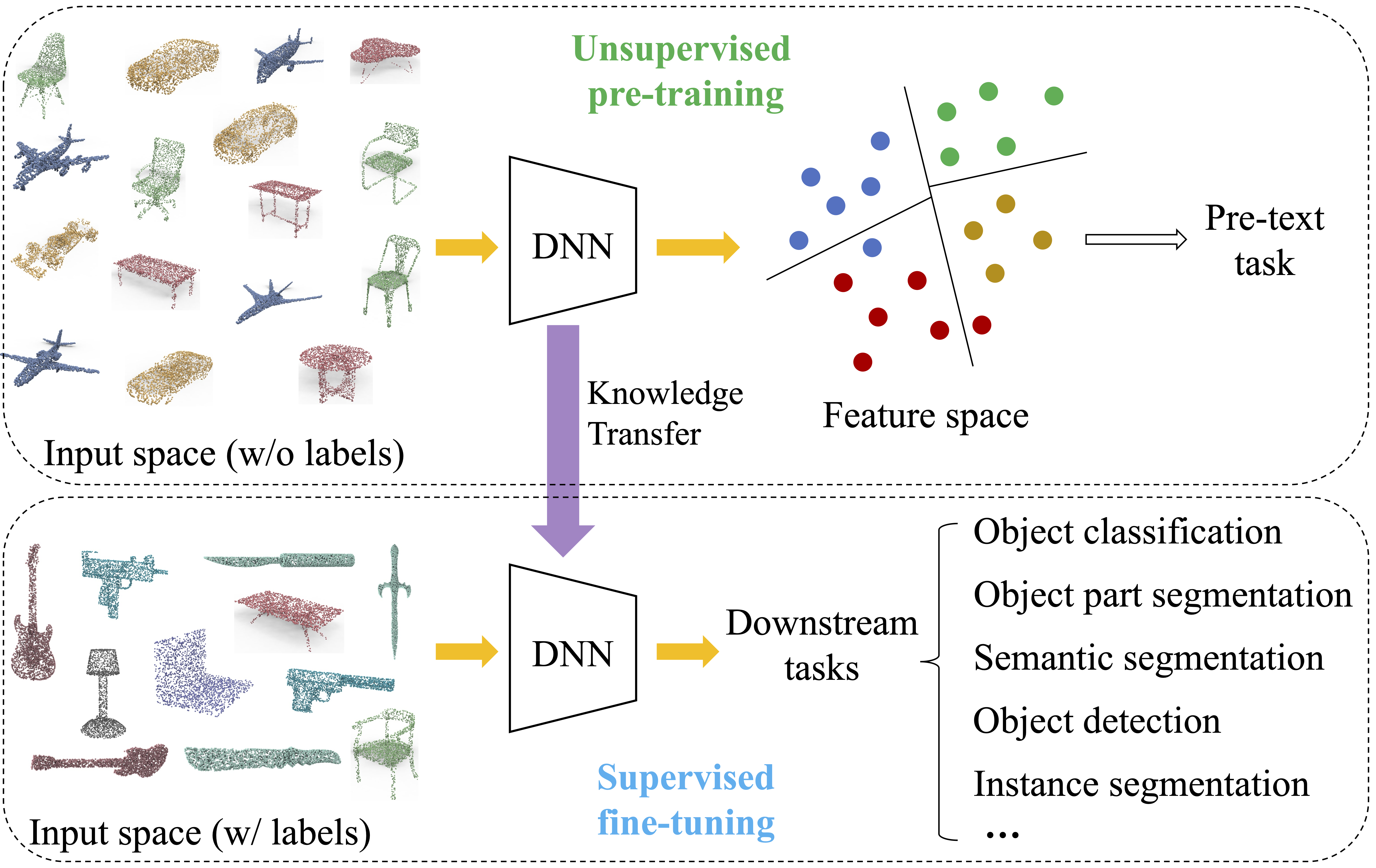}
    \caption{Typical pipeline for self-supervised pertaining. The figure is extracted from \cite{xiao2022unsupervised}.}
    \label{fig:ssl}
    \vspace{-10pt}
\end{figure}

Self-supervised pretraining learns from large-scale unlabelled point clouds, and the learnt parameters can be applied to initialize downstream networks for faster convergence and effective learning from small task data, as illustrated in Fig.~\ref{fig:ssl}. It has attracted increasing interest as it can work with no human annotations. A milestone is PointContrast~\cite{xie2020pointcontrast} which learns network weights from 3D scene frames and fine-tunes networks on multiple high-level 3D tasks such as semantic segmentation and object detection. However, the performance gains of self-supervised pretraining remain limited compared with 2D image and NLP pretraining. 

Most existing studies tackle point cloud pretraining via two approaches: contrastive pretraining and generative pretraining.
Contrastive pretraining~\cite{xie2020pointcontrast,huang2021spatio,chen20224dcontrast, liu2023fac} adopts a discriminative approach and it learns by maximizing the similarity of positive pairs (different augmentations of the same sample or different views of the same scene) while minimizing similarity between negative pairs (different samples). This enhances the network's ability to distinguish between similar and dissimilar examples, leading to improved generalization performance. 
Differently, generative pretraining~\cite{xu2023mv,yang2022gd} learns to generate new point-cloud samples with similar input distribution. The learned model captures representative features of input data which can be fine-tuned for downstream tasks. 
Other studies also combine self-supervised objective with supervised learning tasks to enhance model performance \cite{gadelha2020label}.
Recently, Xiao et al.~\cite{xiao2022unsupervised} performed a comprehensive survey of self-supervised learning for point clouds.

\subsection{Multi-modal pretraining}\label{subsec.multimodal}

Unlike self-supervised pretraining that learns from massive unlabelled data, VLMs are pre-trained with image-text pairs crawled from the Internet. The objective is to train a model to understand the relationships between images and their corresponding textual descriptions.
Its remarkable zero-/few-shot recognition ability has inspired several studies on multi-modal point cloud pretraining. Two typical approaches have been explored: (1) transferring knowledge from existing VLMs to point cloud models and (2) extending the image-text pretraining paradigm for point cloud-text pretraining.
Table~\ref{tab. Sum of PFM methods} summarizes representative multi-model pretraining methods.

\vspace{-5pt}
\subsubsection{From language-vision to point cloud}\label{subsec.multi-modal_1}

VLMs are trained on billions of images with semantic-rich captions. Though they advanced open-vocabulary image understanding tasks greatly, they are not directly applicable in the 3D domain due to the lack of large-scale 3D-text pairs. One line of research aims to leverage the knowledge in VLMs to aid point cloud learning. The primary approach relies on images paired with point clouds as a bridge for knowledge distillation across modalities \cite{lu2023ovir}.

Several pioneering studies~\cite{zhang2022pointclip,huang2023clip2point,zhu2022pointclipv2} transfer CLIP~\cite{radford2021learning} model for \textit{point cloud classification}. For instance, Zhang et al.~\cite{zhang2022pointclip} generate scatter depth maps by projecting raw points onto pre-defined image planes, feed the depth maps to CLIP's visual encoder to extract multi-view features, and obtain zero-shot predictions with a text-generated classifier. 
Some \cite{liu2022partslip,umam2023partdistill} render object-level point clouds into multi-view images of predefined camera poses for \textit{object part segmentation}. The rendered images are fed to the pretrained GLIP\cite{li2022grounded} along with a text prompt for predicting bounding boxes from the text prompt.
Wang et al.~\cite{wang2023vl} exploit visual-linguistic assistance for \textit{3D semantic scene graph prediction}. The method projects point clouds into images and trains a multi-modal model to capture semantics from vision, language, and point clouds, and it adopts CLIP to align the visual-linguistic semantics.

\vspace{-5pt}
\subsubsection{Language-point cloud pretraining}\label{subsec.multi-modal_2}

Inspired by the impressive performance of VLMs, researchers are now exploring the extension of the vision-language pretraining for point-cloud learning. However, collecting Internet-scale point-text samples is extremely difficult. To overcome this challenge, recent studies exploit VLMs to generate captions for image data that can be easily obtained and aligned with point clouds, producing an abundance of point-text pairs for pretraining. This approach allows learning rich and transferrable 3D visual-semantic representations with little human annotations.

For example, Xue et al.~\cite{xue2022ulip} design cross-modal contrastive learning to learn a unified representation of images, texts, and point clouds for \textit{3D shape classification}. They adopt CLIP to generate training triplets to learn a 3D representation space that aligned with the image-text space. Zeng et al.\cite{zeng2023clip2} explore contrastive language-image-point pretraining for \textit{point cloud object recognition}. They adopt DetCLIP\cite{yaodetclip} to extract image proposals given language captions, employ the proposals to parse corresponding point cloud instances, and conduct cross-modal contrastive pretraining to learn semantic-level language-3D alignment between texts and point clouds as well as instance-level image-3D alignment between images and point clouds.

While most studies focus on object-level point cloud understanding, several studies~\cite{ding2022language,yang2023regionplc,lu2023open,yan2024maskclustering,nguyen2023open3dis,yuan2023visual} explore \textit{open-vocabulary scene understanding} with VLM knowledge. They tackle different tasks such as \textit{3D semantic segmentation}, \textit{3D object detection}, and \textit{3D instance segmentation}, aiming to localize and recognize categories that are not present in the annotated label space. For example, Ding et al.\cite{ding2022language}, generate captions for images of 3D indoor scenes to create hierarchical \textit{point-caption} pairs including scene-, view-, and entity-level captions. The pairs provide coarse-to-fine supervision signals and help learn appropriate 3D visual-semantic representations with contrastive learning. With frozen text encoder in BERT\cite{devlin2018bert} or CLIP, category embeddings can be extracted as text-embedded semantic classifier for recognition.

Though \cite{ding2022language} achieves promising results, it is often hindered by its coarse image-level inputs. Consequently, it largely identifies sparse and salient scene objects only, making it difficult for dense understanding tasks such as semantic and instance segmentation. Yang et al.~\cite{yang2023regionplc} address this issue by introducing dense visual prompts that elicit region-level visual-language knowledge via captioning. The method allows creation of dense regional point-language associations, enabling point-discriminative contrastive learning for point-independent learning from captions as well as better open-vocabulary scene understanding.

\subsection{Summary and discussion}\label{subsec.pretrain-summary}

PFMs have shown great potential in enhancing point cloud learning with minimal human annotations. While this field is highly active, it remains underexplored, offering numerous opportunities for further research. Self-supervised pretraining, although effective in 2D computer vision and NLP, has not yet achieved the same impact in point cloud learning, where random initialization still predominates. This is mainly due to the scarcity of large-scale point cloud datasets and the lack of unified, generalizable point cloud backbone models~\cite{xiao2022unsupervised}.

Moreover, the potential of pre-training language-point cloud foundation models remains largely untapped. A significant challenge lies in creating large-scale point-text pairs for pretraining, as collecting vast amounts of 3D data, especially paired with text, is not feasible. While leveraging existing VLMs can help, it still requires gathering numerous point clouds and images, with the images serving as a bridge for knowledge transfer. 
Most studies focus on object-level point clouds or indoor point clouds while few tackles outdoor LiDAR data due to larger difficulties. 
Despite these challenges, this research direction is promising, and further exploration is expected to fully realize its potential.

\section{Evaluation and Benchmarks}\label{sec.benchmarks}

\subsection{Pros and Cons}

As outlined in the preceding description, existing label-efficient learning methods have distinct setups and data prerequisites, each tailored to specific working scopes and application scenarios. In this section, we discuss their respective strengths and limitations.

\noindent \textbf{Data augmentation} can be ubiquitously utilized in training various deep point cloud models, with the advantages of increased data diversity, enhanced generalization, reduced over-fitting, and improved training convergence. However, the effectiveness of data augmentation varies with tasks and datasets. In addition, many data augmentation methods have sophisticated designs and increased computational costs which limits the scope of their applications greatly.

\noindent \textbf{Domain transfer learning}. Domain adaptation allows good utilization of existing annotations while handling new data. It is beneficial for mitigating the demand for target data annotations considering the high cost associated with point cloud labelling. Nevertheless, adapting towards a target domain without explicit target supervision is challenging especially while facing large inter-domain distribution gaps. Unsupervised domain adaptation aims to adapt with unlabelled target data, but it requires complicated tuning with multiple hyperparameters. Domain generalization trains generalizable models that are more robust to data variations across domains. It is very useful for point cloud models that are often deployed in diverse and dynamic environments. However, point clouds from different sources often exhibit great heterogeneity, making it challenging to learn a universal feature representation that works well across domains. Beyond that, domain generalization models are sensitive to drastic domain shifts, where the characteristics of unseen domains significantly differ from those in the training domains.

\noindent \textbf{Weakly-supervised learning} requires much fewer labelled point clouds, which can significantly reduce point-cloud annotation efforts and enable good scalability towards larger datasets and real-world scenarios. In addition, it is flexible in both the types of point-cloud annotations and unlabelled training data. On the other hand, due to the ambiguity and noises in weak labels as well as the weak supervision, it tends to learn biased representations while handling complex perception tasks.

\begin{table}[t]
    \caption{Label efficient semantic segmentation on ScanNet-V2. $^{\dag}$ for active learning. For simplicity, we omit the ‘\%’ after the value. ``*" means reproduced results in corresponding papers.}
    \label{tab:benchmark-semseg-scannet}
    \vspace{-5pt}
    \centering
    \tiny
    \setlength\tabcolsep{4.2pt}
    \begin{tabular}{|l|c|c|c|c|c|c|c|c|}
    \hline
       \textbf{Method} & \textbf{Backbone} & \multicolumn{3}{c}{\textbf{Supervision}} & \multicolumn{2}{|c}{\textbf{mIoU@val}} & \multicolumn{2}{|c|}{\textbf{mIoU@test}}\\
    \hline
       \multicolumn{9}{|c|}{Fully-Supervised Learning} \\
    \cdashline{1-9}
       PointNet++\cite{qi2017pointnet++} & & \multicolumn{3}{c}{100\%} & \multicolumn{2}{|c}{53.5} & \multicolumn{2}{|c|}{33.9}\\
       Rigid KPConv\cite{thomas2019KPConv} & & \multicolumn{3}{c}{100\%} & \multicolumn{2}{|c}{68.5*} & \multicolumn{2}{|c|}{68.6}\\
       Deformable KPConv\cite{thomas2019KPConv} & & \multicolumn{3}{c}{100\%} &  \multicolumn{2}{|c}{69.3*} & \multicolumn{2}{|c|}{68.4} \\
       RandLA-Net\cite{hu2020randla} & & \multicolumn{3}{c}{100\%} & \multicolumn{2}{|c}{} & \multicolumn{2}{|c|}{64.5} \\	
       3D sparse U-Net & & \multicolumn{3}{c}{100\%} & \multicolumn{2}{|c}{72.9} & \multicolumn{2}{|c|}{} \\
       MinkowskiNet\cite{choy20194d} & & \multicolumn{3}{c}{100\%} & \multicolumn{2}{|c}{72.2} & \multicolumn{2}{|c|}{73.4}  \\
       PCAM\cite{wei2020multi} & & \multicolumn{3}{c}{100\%} & \multicolumn{2}{|c}{} & \multicolumn{2}{|c|}{} \\
       MIL\cite{yang2022mil} & & \multicolumn{3}{c}{100\%} & \multicolumn{2}{|c}{73.3} & \multicolumn{2}{|c|}{}  \\
    \hline
       \multicolumn{9}{|c|}{Data Augmentation} \\
    \cdashline{1-9}
       \multirow{3}{8em}{Mix3D\cite{nekrasov2021mix3d}} & MinkowskiNet & \multicolumn{3}{c}{100\%} & \multicolumn{2}{|c}{72.4*/73.6} & \multicolumn{2}{|c|}{78.1} \\
       & Rigid KPConv & \multicolumn{3}{c}{100\%} & \multicolumn{2}{|c}{68.8*/69.5} & \multicolumn{2}{|c|}{}\\
       & Deformable KPConv & \multicolumn{3}{c}{100\%} & \multicolumn{2}{|c}{69.3*/70.3} & \multicolumn{2}{|c|}{} \\
    \hline
       \multicolumn{9}{|c|}{Weakly-Supervised Learning} \\
    \cdashline{1-9}
       MPRM\cite{wei2020multi} & PCAM & \multicolumn{3}{c}{subcloud-level} & \multicolumn{2}{|c}{43.2} & \multicolumn{2}{|c|}{41.1} \\
       MIL\cite{yang2022mil} & MIL & \multicolumn{3}{c}{subcloud-level} & \multicolumn{2}{|c}{47.4} & \multicolumn{2}{|c|}{45.8} \\
    \cdashline{1-9}
       WyPR\cite{ren20213d} & PointNet++ & \multicolumn{3}{c}{scene-level} & \multicolumn{2}{|c}{29.6} & \multicolumn{2}{|c|}{24.0} \\
       MIL\cite{yang2022mil} & MIL & \multicolumn{3}{c}{scene-level} & \multicolumn{2}{|c}{26.2} & \multicolumn{2}{|c|}{} \\
    \cdashline{1-9}
       1T1C\cite{liu2021one} & 3D sparse U-Net & \multicolumn{3}{c}{0.02\%} & \multicolumn{2}{|c}{70.5} & \multicolumn{2}{|c|}{69.1}  \\
    \cdashline{1-9}[0.8pt/2pt]
       \multirow{2}{8em}{Zhang et al.\cite{zhang2021weakly}} & \multirow{2}{6em}{RandLA-Net} & \multicolumn{3}{c}{1\%} & \multicolumn{2}{|c}{} & \multicolumn{2}{|c|}{51.1} \\
       & & \multicolumn{3}{c}{10\%} & \multicolumn{2}{|c}{} & \multicolumn{2}{|c|}{52.0}\\
    \cdashline{1-9}[0.8pt/2pt]
       PSD\cite{zhang2021perturbed} & 	RandLA-Net & \multicolumn{3}{c}{1\%} & \multicolumn{2}{|c}{} & \multicolumn{2}{|c|}{50.3/54.7} \\
       HybridCR\cite{li2022hybridcr} & RandLA-Net & \multicolumn{3}{c}{1\%} & \multicolumn{2}{|c}{56.9} & \multicolumn{2}{|c|}{56.8} \\
       GaIA\cite{zhang2021weakly}  & 3D sparse U-Net & \multicolumn{3}{c}{1\%} & \multicolumn{2}{|c}{} & \multicolumn{2}{|c}{65.2} \\
       SQN\cite{hu2022sqn} & RandLA-Net & \multicolumn{3}{c}{0.1\%} & \multicolumn{2}{|c}{} & \multicolumn{2}{|c|}{56.9}\\
       CPCM\cite{liu2023cpcm} & 3D sparse U-Net & \multicolumn{3}{c}{0.1\%} & \multicolumn{2}{|c}{63.8} & \multicolumn{2}{|c|}{62.5} \\ 
       HPAL$^{\dag}$\cite{xu2023hierarchical} & MinkowskiNet & \multicolumn{3}{c}{0.1\%} & \multicolumn{2}{|c}{69.9} & \multicolumn{2}{|c|}{68.2} \\
    \cdashline{1-9}
       GaIA\cite{zhang2021weakly} & 3D sparse U-Net & \multicolumn{3}{c}{20 pts/scene} & \multicolumn{2}{|c}{} & \multicolumn{2}{|c|}{63.8} \\
       DAT\cite{wu2022dual} & Rigid KPConv & \multicolumn{3}{c}{20 pts/scene} & \multicolumn{2}{|c}{54.6/58.9} & \multicolumn{2}{|c|}{51.6/55.2} \\
       MIL\cite{yang2022mil} & MIL & \multicolumn{3}{c}{20 pts/scene} & \multicolumn{2}{|c}{57.8} & \multicolumn{2}{|c|}{54.4}\\
       CPCM\cite{liu2023cpcm} & 3D sparse U-Net & \multicolumn{3}{c}{20 pts/scene} & \multicolumn{2}{|c}{62.7} & \multicolumn{2}{|c|}{62.8}\\
       HPAL$^{\dag}$\cite{xu2023hierarchical} & MinkowskiNet & \multicolumn{3}{c}{20 pts/scene} & \multicolumn{2}{|c}{62.2} & \multicolumn{2}{|c|}{62.5}\\
    \cdashline{1-9}
       \multicolumn{2}{|c|}{mIoU@val} & 1\% & 5\% & 10\% & 20\% & 30\% & 40\% & 100\% \\
    \cdashline{1-9}[0.8pt/2pt]
       3D sparse U-Net & \multirow{3}{8em}{3D sparse U-Net} & 40.9 & 48.1 & 57.2 & 64.0 & 67.1 & 68.8 & 72.9 \\
       GPCL\cite{jiang2021guided} & & & 54.8 & 60.5 & 66.7 & 68.9 & 71.3 & 74.0 \\
       WS3D\cite{liu2022weakly} &  & 49.9 & 56.2 & 62.2 & 69.4 & 70.3 & 73.4 & 76.9 \\
    \hline
    \end{tabular}
\end{table}

\begin{table}[t]
    \caption{Label efficient semantic segmentation on SemanticKITTI. $^{\dag}$ denotes active training. For simplicity, we omit ‘\%’ after the mIoU numbers. ``src" denotes source domain. Two sampling strategies (abbr. 'SPL') for partitioning labelled and unlabelled training data under Weakly-Supervised Learning: 'SS' for sequential sampling, 'US' for uniform sampling.}
    \label{tab:benchmark-semseg-semantickitti}
    \vspace{-5pt}
    \centering
    \tiny
    \setlength\tabcolsep{4pt}
    \begin{tabular}{|l|c|c|c|c|c|c|c|c|c|c|}
    \hline
       \textbf{Method} & \textbf{Backbone} & \multicolumn{2}{c}{\textbf{Supervision}} & \multicolumn{3}{|c|}{\textbf{mIoU@val}} & \multicolumn{3}{|c|}{\textbf{mIoU@test}} \\
    \hline
       \multicolumn{10}{|c|}{Fully-Supervised Learning} \\
    \cdashline{1-10}
       RandLA-Net\cite{hu2020randla} & & \multicolumn{2}{c|}{100\%} & \multicolumn{3}{c|}{} & \multicolumn{3}{c|}{53.9} \\	
       MinkowskiNet\cite{choy20194d} & & \multicolumn{2}{c|}{100\%} & \multicolumn{3}{c|}{58.9} & \multicolumn{3}{c|}{}\\
       SPVCNN\cite{tang2020searching} & & \multicolumn{2}{c|}{100\%} & \multicolumn{3}{c|}{60.7} & \multicolumn{3}{c|}{} \\
       Cylinder3D\cite{zhu2021cylindrical} & & \multicolumn{2}{c|}{100\%} & \multicolumn{3}{c|}{64.3} & \multicolumn{3}{c|}{68.9} \\
    \hline
       \multicolumn{10}{|c|}{Data Augmentation} \\
    \cdashline{1-10}
       \multirow{2}{8em}{PolarMix\cite{xiao2022polarmix}} & MinkowskiNet & \multicolumn{2}{c}{100\%} & \multicolumn{3}{|c}{58.9/65.0} & \multicolumn{3}{|c|}{} \\
       & SPVCNN & \multicolumn{2}{c}{100\%} & \multicolumn{3}{|c}{60.7/66.2} & \multicolumn{3}{|c|}{} \\
    \hline
       \multicolumn{10}{|c|}{Unsupervised Domain Adaptation} \\
    \cdashline{1-10}
       PCT\cite{xiao2022transfer} & MinkowskiNet & \multicolumn{2}{c}{source: synlidar} & \multicolumn{3}{|c}{28.9} & \multicolumn{3}{|c|}{}\\
       PolarMix\cite{xiao2022polarmix} & MinkowskiNet & \multicolumn{2}{c}{source: synlidar} & \multicolumn{3}{|c}{31.0} & \multicolumn{3}{|c|}{}\\
       CoSMix\cite{saltori2022cosmix} & MinkowskiNet & \multicolumn{2}{c}{source: synlidar} & \multicolumn{3}{|c}{32.2} & \multicolumn{3}{|c|}{}\\
       SCT\cite{xiao2023domain} & MinkowskiNet & \multicolumn{2}{c}{source: synlidar} & \multicolumn{3}{|c}{36.0} & \multicolumn{3}{|c|}{}\\
       Annotator$^{\dag}$\cite{xie2023annotator} & MinkowskiNet & \multicolumn{2}{c}{source: synlidar} & \multicolumn{3}{|c}{53.7} & \multicolumn{3}{|c|}{}\\
       DGT-ST\cite{yuan2024density} & MinkowskiNet & \multicolumn{2}{c}{source: synlidar} & \multicolumn{3}{|c}{43.1} & \multicolumn{3}{|c|}{}\\
    \hline
       \multicolumn{10}{|c|}{Weakly-Supervised Learning} \\
    \cdashline{1-10}
       \multirow{3}{8em}{Scribble\cite{unal2022scribble}} &  Cylinder3D & \multicolumn{2}{c}{scribble} & \multicolumn{3}{|c}{61.3} & \multicolumn{3}{|c|}{} \\ 
       & MinkowskiNet & \multicolumn{2}{c}{scribble} & \multicolumn{3}{|c}{58.5} & \multicolumn{3}{|c|}{}\\
       & SPVCNN & \multicolumn{2}{c}{scribble} & \multicolumn{3}{|c}{60.8} & \multicolumn{3}{|c|}{}\\
    \cdashline{1-10}
       HybridCR\cite{li2022hybridcr} & RandLA-Net & \multicolumn{2}{c}{1\%} & \multicolumn{3}{|c}{51.9} & \multicolumn{3}{|c|}{52.3}\\
    \cdashline{1-10}[0.8pt/2pt]
       \multirow{2}{8em}{SQN\cite{hu2022sqn}} & \multirow{2}{5.8em}{RandLA-Net} & \multicolumn{2}{c}{0.1\%} & \multicolumn{3}{|c}{} & \multicolumn{3}{|c|}{50.8}\\
       & & \multicolumn{2}{c}{0.01\%} & \multicolumn{3}{|c}{} & \multicolumn{3}{|c|}{39.1} \\
    \cdashline{1-10}[0.8pt/2pt]
       \multirow{2}{8em}{LESS\cite{liu2022less}} & \multirow{2}{5em}{Cylinder3D} & \multicolumn{2}{c}{0.1\%} & \multicolumn{3}{|c}{66.0} & \multicolumn{3}{|c|}{} \\ 
       & & \multicolumn{2}{c}{0.01\%} & \multicolumn{3}{|c}{61.0} &  \multicolumn{3}{|c|}{} \\
    \cdashline{1-10}
       \multicolumn{2}{|c|}{mIoU@val} & 1\% & 5\% & 10\% & 20\% & 30\% & 40\% & 100\% & SPL \\
    \cdashline{1-10}[0.8pt/2pt]
       3D sparse U-Net & \multirow{3}{7.5em}{3D sparse U-Net} & 28.6 & 34.8 & 43.9 & 53.8 & 55.4 & 57.4 & 65.0 & -\\
       GPCL\cite{jiang2021guided} & & & 41.8 & 49.9 & 58.8 & 59.4 & 59.9 & 65.8  & SS\\
       WS3D\cite{liu2022weakly} &  & 38.9 & 43.7 & 52.3 & 61.4 & 62.1 & 63.2 & 66.9 & SS\\
    \cdashline{1-10}[0.8pt/2pt]
       LaserMix\cite{kong2022lasermix} &  Cylinder3D &  & 	56.7 & 60.0 & 61.9 & 62.1 & 62.3 &  & US \\
       LiM3D\cite{li2023less} &  Cylinder3D & 58.4 & 59.5 & 62.2 & 63.1 & 63.3 & 63.6 & 69.5 & US\\
    \cdashline{1-10}
       \multicolumn{2}{|c|}{mIoU@val} & 1\% & 2\% & \multicolumn{2}{c|}{3\%} & \multicolumn{2}{c|}{4\%} & \multicolumn{2}{c|}{5\%} \\
    \cdashline{1-10}[0.8pt/2pt]
       ReDAL$^{\dag}$\cite{wu2021redal} & MinkowskiNet & 37.5 & 48.9 & \multicolumn{2}{c|}{55.3} & \multicolumn{2}{c|}{58.4} & \multicolumn{2}{c|}{59.8} \\
       & SPVCNN & 41.9 & 51.7 & \multicolumn{2}{c|}{55.8} & \multicolumn{2}{c|}{56.9} & \multicolumn{2}{c|}{58.2}\\
    \cdashline{1-8}[0.8pt/2pt]
       LiDAL$^{\dag}$\cite{hu2022lidal} & MinkowskiNet & 47.3 & 56.7 & \multicolumn{2}{c|}{58.7} & \multicolumn{2}{c|}{59.5} & \multicolumn{2}{c|}{60.1} \\
       & SPVCNN & 48.8 & 57.1 & \multicolumn{2}{c|}{58.7} & \multicolumn{2}{c|}{59.3} & \multicolumn{2}{c|}{59.5} \\
    \hline
    \end{tabular}
\end{table}

\noindent \textbf{Pre-trained foundation models} have been explored for label-efficient learning with point clouds. Specifically, self-supervised learning allows learning effective representations from unlabelled point clouds, thereby alleviating the demand for large-scale annotations for downstream tasks. Multi-modal pre-training also demonstrates great potential in generalization and zero-shot transfer. Nonetheless, these studies require substantial and diverse point clouds in training, presenting notable challenges for both research and practical implementation. The development of pre-trained point-cloud foundation models remains at an exploratory stage, highlighting the necessity for further development in this fast-evolving research area.

\subsection{Performance}\label{subsec.performance}

In this section, we conduct a comprehensive analysis of representative 3D label-efficient learning methods and benchmark them with fully supervised learning models to assess their effectiveness in reducing annotation costs. Our evaluation encompasses widely used benchmark suites, such as ScanNet-V2 for indoor point clouds and KITTI/SemanticKITTI for outdoor environments. The benchmarking focuses on 3D object detection and 3D semantic segmentation, where all experiments are evaluated with official criteria of fully supervised learning benchmarks for fairness. All performance metrics are sourced directly from the respective papers as well.

It's important to note that label-efficient learning covers a wide range of perception tasks and learning setups. For fairness in comparison, we exclude methods evaluated with modified or customized benchmarks for specific purposes. For example, many UDA and DG studies use shared classes between source and target domains, reducing the class number compared to fully supervised learning. Additionally, some tasks, such as 3D instance segmentation, are not included in the benchmarking due to very limited research. For more detailed benchmarking of these methods, please refer to the original papers.

Tables~\ref{tab:benchmark-semseg-scannet} and \ref{tab:benchmark-semseg-semantickitti} provide a comprehensive overview of the performance of label-efficient \textit{semantic segmentation} on the indoor dataset ScanNet-V2 and the outdoor dataset SemanticKITTI, respectively. The evaluation metric is based on the official criterion of mean Intersection over Union (mIoU). In addition, Tables~\ref{tab:benchmark-det-scannet} and \ref{tab:benchmark-det-kitti} present the results of label-efficient \textit{object detection} on the ScanNet-V2 and KITTI benchmarks, respectively. Recognizing the pivotal role of backbone models, we include them in the tables to facilitate a more robust comparison. Notably, we only listed the backbones that have been frequently adopted in label-efficient learning for brevity and relevance.

It is inspiring to witness the remarkable capacity of label-efficient learning in substantially reducing the need for costly annotations of point-cloud data. With much fewer annotations, the performance of some reported methods is even on par with that of fully supervised methods. This underscores the immense potential of this research direction, and we anticipate a surge in further exploration and advancements in this promising research area.

\begin{table}[t]
    \caption{Label efficient object detection on the validation set of ScanNet-V2. ``@0.25" and ``@0.5" mean mAP@0.25 and mAP@0.5, respectively. For simplicity, we omit the ‘\%’ after the value.}
    \label{tab:benchmark-det-scannet}
    \vspace{-5pt}
    \centering
    \tiny
    \setlength\tabcolsep{5pt}
    \begin{tabular}{|l|c|cc|cc|cc|cc|}
    \hline
        \multirow{2}{4em}{Method} & \multirow{2}{5em}{Backbone} & \multicolumn{2}{c|}{5\%} & \multicolumn{2}{c|}{10\%} & \multicolumn{2}{c|}{20\%} & \multicolumn{2}{c|}{100\%}\\
        & & @0.25 & @0.5 & @0.25 & @0.5 & @0.25 & @0.5 & @0.25 & @0.5\\
    \hline
        VoteNet\cite{qi2019deep} & & & 31.0 & 11.9 & 41.6 & 21.2 & 58.6 & 33.5 &\\
    \cdashline{1-10}
        SESS\cite{zhao2020sess} & VoteNet & & & 39.7 & 18.6 & 47.9 & 26.9 & 62.1 & 38.8 \\
        3DIoUMatch\cite{wang20213dioumatch} & VoteNet & 40.0 & 22.5 & 47.2 & 28.3 & 52.8 & 35.2 & 62.9 & 42.1 \\
        Wang et al.\cite{wang2023not} & VoteNet & 40.5 & 23.8 & 48.8 & 31.15 & 4.5 & 37.3 & 63.8 & 44.1 \\
        DQS3D\cite{gao2023dqs3d} & VoteNet & 53.2 & 35.6 & 55.7 & 38.2 & 58.0 & 42.3 & 64.1 & 48.2 \\
    \cdashline{1-10}[0.8pt/2pt]
        BR\cite{xu2022back} & VoteNet & \multicolumn{6}{l|}{position-level annotations} & 35.5 & \\
    \hline
    \end{tabular}
\end{table}

\begin{table}[t]
    \caption{Label efficient object detection on KITTI. We report AP\textsuperscript{3D} on \textit{val} set. 3D bounding box IoU threshold is 0.7 for cars and 0.5 for pedestrians and cyclists ‘E’, ‘M’ and ‘H’ represent easy, moderate and hard classes of objects, respectively.``src" denotes source domain. For simplicity, we omit the ‘\%’ after the value.}
    \label{tab:benchmark-det-kitti}
    \vspace{-5pt}
    \centering
    \tiny
    \setlength\tabcolsep{2.0pt}
    \begin{tabular}{|l|l|l|ccc|ccc|ccc|l|}
    \hline
        \multirow{2}{4em}{Method} & \multirow{2}{5em}{Backbone} & \multirow{2}{5.5em}{Supervision} & \multicolumn{3}{c|}{Car} & \multicolumn{3}{c|}{Pedestrian} & \multicolumn{3}{c|}{Cyclist} & \multirow{2}{0.5em}{Set}\\
        & & & E & M & H & E & M & H & E & M & H & \\
    \hline
        \multicolumn{13}{|c|}{Supervised Learning} \\
    \cdashline{1-13}
       StarNet\cite{ngiam2019starnet} &  & 100\%  & 81.6 & 74.0 & 67.1 & 48.6 & 41.3 & 39.7 & 73.1 & 58.3 & 52.6 & test\\
       SECOND\cite{yan2018second} & & 100\% & 83.3 & 72.5 & 65.8 & 47.0 & 38.8 & 34.9 & 71.3 & 52.1 & 45.8 & test \\
       PV-RCNN\cite{shi2020pv-rcnn} & & 100\% & 90.3 & 81.4 & 76.8 & 52.2 & 43.3 & 40.3 & 78.6 & 63.7 & 57.7 & test\\
       PointRCNN\cite{shi2019pointrcnn} & & 100\% & 87.0 & 75.6 & 70.7 & 48.0 & 39.4 & 36.0 & 75.0 & 58.8 & 52.5 & test\\
       Voxel-RCNN\cite{deng2021voxel} & & 100\% & 90.9 & 81.6 & 77.1 & & & & & & & test\\
    \hline
       \multicolumn{13}{|c|}{Data Augmentation} \\
    \cdashline{1-13}
        PPBA\cite{cheng2020improving} & StarNet & 100\% & 84.2 & 77.7 & 71.2 & 52.7 & 44.1 & 41.5 & 79.4 & 62.0 & 55.3 & test\\
    \hline
       \multicolumn{13}{|c|}{Unsupervised Domain Adaptation} \\
    \cdashline{1-13}
        \multirow{2}{4em}{SN\cite{wang2020train}} & \multirow{2}{4em}{PointRCNN} & src:nuScenes & 13.2 & 12.1 & 11.1 & & & & & & & val \\
        & & src:Waymo & 13.1 & 14.9 & 14.4 & & & & & & & \\
    \cdashline{1-13}[0.8pt/2pt]
        CDN & PointRCNN & src:PreSIL & & 19.0 & & & 13.2 & & & 9.1 & & test \\
    \cdashline{1-13}[0.8pt/2pt]
        \multirow{4}{4em}{ST3D\cite{yang2021st3d}} & SECOND & src:Waymo & & 73.4 & & & & & & & & \multirow{4}{1.5em}{val}\\
        & PV-RCNN  & src:Waymo & & 76.9 & & & & & & & & \\
        & SECOND  & src:nuScenes & & 62.6 & & & & & & & & \\
        & PV-RCNN  & src:nuScenes & & 72.9 & & & & & & & & \\
    \cdashline{1-13}[0.8pt/2pt]
        SRDAN\cite{zhang2021srdan} & SECOND & src:PreSIL & 25.9 & 22.1 & 18.7 & 15.9 & 14.6 & 12.5 & 9.6 & 9.4 & 9.1 & val\\
    \cdashline{1-13}[0.8pt/2pt]
        \multirow{2}{4em}{MLC-Net\cite{luo2021unsupervised}} & \multirow{2}{4em}{PointRCNN} & src:Waymo & 69.4 & 59.4 & 56.3 & & & & & & & \multirow{2}{1.5em}{val} \\
        & & src:nuScenes & 71.3 & 55.4 & 49.0 & & & & & & &\\
    \cdashline{1-13}[0.8pt/2pt]
        \multirow{2}{4em}{SPG\cite{xu2021spg}} & PointPillars & \multirow{2}{4em}{src:Waymo} & 89.8 & 81.4 & 78.9 & 59.7 & 53.6 & 49.2 & 83.3 & 66.1 & 62.0 & \multirow{2}{1.5em}{val} \\
        & PV-RCNN & & 92.5 & 85.3 & 82.8 & 69.7 & 61.8 & 56.4 & 91.8 & 74.4 & 69.5 & \\
    \cdashline{1-13}[0.8pt/2pt]
        \multirow{3}{5.5em}{REDB\cite{chen2023revisiting}} & SECOND  & src:Waymo &  & 54.1 & 52.6 &  & 48.2 & 43.2 &  & 48.0 & 47.2 & \multirow{3}{1.5em}{val} \\
        & SECOND  & src:nuScenes & & 51.3 & 46.2 & 18.4 & 18.9 &  & 26.1 & 27.3 & & \\
        & PointRCNN & src:nuScenes & 71.5 & 57.9 & 53.9 & 52.3 & 44.3 & 38.0 & 45.1 & 32.9 & 31.1 & \\
    \hline
       \multicolumn{13}{|c|}{Weakly-Supervised Learning} \\
    \cdashline{1-13} 
       WS3D\cite{meng2020weakly} & WS3D & Scenes+instances & 84.1 & 75.1 & 73.3 & 74.7 & 70.0 & 66.5 &  &  & & \\
    \cdashline{1-13}[0.8pt/2pt]
       \multirow{2}{5.9em}{HSSDA\cite{liu2023hierarchical}} & \multirow{2}{6em}{Voxel-RCNN} & 1\% & 92.5 & 81.7 &  77.5 & 50.7 & 43.9 & 42.4 & 65.2 & 48.3 & 42.5 & \multirow{2}{1.5em}{val} \\
       & & 2\% & 91.6 & 82.0 & 77.9 & 64.9 & 58.3 & 50.9 & 88.0 & 65.7 & 60.9 & \\
    \cdashline{1-13}[0.8pt/2pt]
        \multirow{5}{5em}{SS3D\cite{liu2022ss3d}} & PointRCNN & \multirow{3}{5em}{Sparse (20\%)} & 87.2 & 77.1 & 76.1 & & & & 86.6 & 73.2 & 66.9 & \multirow{5}{1.5em}{val} \\
        &  PV-RCNN & & 89.5 & 79.3 &  78.3 & & & & 88.0 & 70.4 & 67.4  & \\
        & Voxel-RCNN & & 89.3 & 84.3 & 78.2 & & & & & & & \\
    \cdashline{2-12}[0.8pt/2pt]
        & PV-RCNN & 1\% & 96.2 & 88.1 & 86.9 & 61.7 & 58.7 & 54.5 & 85.6 & 62.8 & 58.4 & \\
        & PV-RCNN & 2\% & 98.3 & 89.2 & 88.3 & 67.5 & 62.3 & 61.0 & 90.1 & 72.2 & 68.3 & \\
    \hline
    \end{tabular}
    \vspace{-5pt}
\end{table}

\section{Future Directions}\label{sec.future}
Label-efficient learning for point clouds remains a very challenging and open research task. In this section, we share our insights on future research directions in label-efficient point cloud learning, pinpointing what are missing in the current research and what are worth further exploration. Specifically, we discuss potential research directions from a general perspective, including \textit{data challenges} in Sec. \ref{sec.future-data_challenge}, \textit{model architectures} in Sec. \ref{sec.future-models}, and each specific label-efficient learning branches in Sec. \ref{sec.future-algorithms}.

\subsection{Data Challenges}\label{sec.future-data_challenge}

\noindent\textbf{Efficient labelling pipelines and tools}. Point cloud annotation is far more laborious than annotating images which largely explains the scarcity of large-scale point cloud datasets. Efficient annotation tools and automatic/semi-automatic annotation techniques have been attempted but their performance cannot meet the increasing demand of large-scale point cloud data. More efficient annotation tools and techniques are urgently needed for better exploitation of the very useful point cloud data.

\noindent \noindent\textbf{Next-generation datasets.} 
Most existing point-cloud datasets have limited scales and diversity as listed in Table \ref{tab:datasets}. The performance over them tends to saturate with prevalent deep networks under a supervised setup, hampering further research in robustness and generalization assessment. Under such circumstances, collecting significantly larger and more diverse datasets is crucial for advancing point cloud related studies. This is well aligned with the recent advancement in PFM, where self-supervised pretraining or multi-modal PFMs necessitate a vast amount point-cloud data in training. Some pioneer efforts such as the Objaverse-XL \cite{deitke2023objaverse} with over 10 million 3D objects represent strides in this direction. However, further endeavours are imperative to cultivate more generalized and uniform 3D representation spaces for facilitating various downstream tasks and applications \cite{zhou2024uni3d}.

\noindent \textbf{Generation for 3D recognition}. 
The progression of visual intelligence is inextricably linked to the availability of large-scale and diverse training data. At the other end, the field of generative AI has unlocked the potential of fabricating synthetic data that closely mimics real-world scenarios \cite{yang2023ai}. In contrast to real data, AI-generated data presents remarkable advantages such as unparalleled abundance, superb scalability, rapid generation, and facile simulation of corner cases. Despite all these enormous potentials, effective utilization of AI-generated 3D point clouds remains largely under-explored while striving to develop robust and accurate 3D perception models.

\subsection{Model Architecture}\label{sec.future-models}

\noindent\textbf{Uniformed architectures.}
Unified backbone structures facilitate knowledge transfer across datasets and tasks which are instrumental to the success of prior deep learning research~\cite{he2016deep, vaswani2017attention}. However, existing work on label-efficient point cloud learning employs very different deep architectures, which poses great challenges for benchmarking and integration as well as benefiting from PFMs. Designing highly efficient and unified deep architectures is a pressing issue with immense value for point cloud learning.

\noindent\textbf{Label-efficient architectures.} Another interesting research direction is label-efficient deep architectures that can achieve competitive performance with much fewer annotations. Several pioneering studies have been conducted, e.g., ~\cite{tang2020searching} for constructing light architectures via neural architecture search, ~\cite{hu2022sqn} for saving annotations by fully exploiting strong local semantic homogeneity of point neighbours, etc.

\subsection{Label-efficient Learning Algorithms}\label{sec.future-algorithms}

Label-efficient learning has been rapidly evolving along different directions in data augmentation in Sec. \ref{sec:data_aug}, domain transfer learning in Sec. \ref{sec.domain_transfer}, weakly-supervised learning in Sec. \ref{sec.weakly}, and pretrained foundation models in Sec. \ref{sec.pretraining}. Nevertheless, research in these areas remains limited as witnessed by the very few papers reviewed in this survey as well as the very low performance in various label-efficient learning benchmarks as listed in Sec. \ref{sec.benchmarks}. At the other end, these gaps also pose great opportunities for future exploration. The subsequent part of this section discusses several potential avenues for future exploration.

\noindent \textbf{Data augmentation} for point cloud recognition could be further explored from four perspectives. The first is to design synthesis networks that can generate extensive and diverse point clouds that can help learn the underlying 3D structure. The second is to design dynamic and adaptive augmentation techniques that allow adjusting the level and type of augmentation according to the model performance and complexity of input data. It can also help better focus on challenging samples, thereby improving the model's robustness. The third is to design task-specific augmentation that is tailored to specific 3D perception and recognition tasks. The fourth is to design physics-aware augmentation that ensures good alignment between the augmented samples and the physical plausibility of 3D sensors.

\noindent \textbf{Domain transfer learning} can be investigated from several perspectives. The first is to develop \textit{model-agnostic} transfer algorithms since point cloud recognition is rapidly evolving with numerous newly proposed deep architectures.
Such algorithms accommodate diverse deep 3D frameworks and facilitate fair evaluations across different backbones as well. 
The second is semi-supervised domain transfer including semi-supervised domain adaptation (SSDA) and semi-supervised domain generalization (SSDG)~\cite{zhou2022domain}. Compared to UDA, SSDA allows accessing a limited number of target annotations for better adaptation in low annotating costs, while in SSDG, only partial source training data are labelled with much reduced annotation budget, both leading to more realistic and practical settings. The third, as illustrated in Sec. \ref{subsec.performance}, is maintaining consistency in training and inference, which is essential for fair benchmarking in domain adaptation and generalization, given the substantial impact that variations in backbone models and hyperparameters could have on performance.

Future research and applications of domain transfer learning could also benefit from incremental learning and online adaptation, which aim to adapt 3D point cloud recognition models toward new domains as relevant data emerges. The online adaptation can also facilitate continuous knowledge updates without retraining on the entire dataset. In addition, addressing the \textit{Sim-to-Real Transfer} gap is essential for real-world applications, requiring techniques that can effectively transfer knowledge from synthetic to real-world point clouds. Furthermore, \textit{multi-modal fusion} incorporates information from diverse modalities such as RGB images, depth maps, sensor data, etc., enhancing the robustness and generalization of point cloud recognition models across domains. Lastly, multi-domain adaptation/generalization maximizes the use of existing data sources for learning robust 3D representations against various perturbations.

\noindent \textbf{Weakly-supervised learning} encompasses diverse setups with varying data prerequisites and configurations, offering multiple avenues for future exploration. For \textit{incomplete supervision}, consistency learning remains underexplored in point cloud recognition though it has shown great potential in general semi-supervised learning~\cite{sohn2020fixmatch,yang2022survey}. Meanwhile, active learning, which intelligently selects informative samples for labelling, is promising by joint utilization of both labelled and unlabelled point clouds. For few-shot learning, we can expect more research on meta-learning on 3D point clouds, together with effective attention mechanisms and data augmentation strategies for more generalizable models under minimal labelled training samples. For \textit{inexact supervision}, multi-task learning has been attracting increasing attention due to the cruciality of understanding relationships among tasks in training. It could be tackled by developing adaptive task weighting mechanisms or exploring inter-task transfer for optimal joint learning. Under the context of open-set learning, increasing research focuses on better model robustness to outliers and anomalies, as well as dynamic adaptation for prompt handling of novel classes. On top of the above listed, learning under \textit{inaccurate supervision} remains sparse for point clouds, necessitating more research in this underexplored area \cite{song2022learning}.

\noindent \textbf{Pre-trained foundation models}. Self-supervised pre-training has shown promising progress in 3D point cloud recognition. However, the potential is far underexplored, and we can expect more advanced 3D self-learning pre-tasks that are tailored for point-cloud data. In addition, we can expect increasing demand for generalized representations that can effectively accommodate the complexity of various downstream tasks in diverse scenarios. Please refer to \cite{xiao2022unsupervised} for more discussion. As for multi-modal pre-training, increasing interest has been observed in how to transfer linguistic and visual knowledge from existing foundation models to point cloud space in an effective, elegant, and resource-saving manner. Furthermore, it remains a formidable task to align feature representations of different types of point clouds (e.g., photogrammetry, LiDAR, and mesh point clouds each of which possesses varying dimensions and information) within a unified foundation model. The ongoing research is striving to unlock the potential benefits of various types of point clouds, presenting both challenges and opportunities in the pursuit of more robust and versatile self-supervised pre-training models.

\section{Conclusion}
Label-efficient learning of point clouds has been investigated extensively over the past decade, leading to plenty of work across different tasks. This survey presents three key points that are critical to research in this field. Firstly, we share the importance and urgency of label-efficient learning in point cloud processing, especially under the context of big data and resource constraints. Secondly, we review four representative label-efficient learning approaches, including data augmentation, domain transfer learning, weakly-supervised learning, and pretrained foundation models, as well as related studies that have achieved very promising outcomes but still have vast space for improvements. Lastly, we comprehensively discuss the progress made in this field and share the challenges and promising future research directions. We expect that this timely and up-to-date survey will inspire more useful studies to further advance this very meaningful research field.

\vspace{-10pt}


{\small
\bibliographystyle{IEEEtran}
\bibliography{ref_arxiv}
}

\appendix[Key concepts]





\noindent {\textbf{Point cloud.} A point cloud is a collection of 3D points, represented by their spatial coordinates in x, y, and z. Depending on the type of point clouds, additional attributes may also be included, e.g., normal values for object-level point clouds~\cite{chang2015shapenet}, color information for indoor dense point clouds~\cite{dai2017scannet}, or intensity value for LiDAR point clouds~\cite{behley2019semantickitti}.}

\noindent {\textbf{Supervised learning} optimizes machine learning models under the full supervision of labels where models learn to map input data to output label space. The training data consists of pairs of input point clouds and corresponding labels, where the labels annotated by humans are exactly the ground truth of the models' output.}

\noindent\textbf{Label-efficient learning} focuses on developing methods that can learn from a limited amount of labeled data. The goal is to reduce the amount of labeled data in deep network training, as labelling data can be time-consuming and expensive.

\noindent\textbf{3D shape classification} aims to identify the category of an object point cloud, such as chairs, tables, cars, and buildings. Categorical labels are needed as ground truth for training 3D classification models. Accuracy, defined as the ratio of correctly classified objects to the total number of objects in the dataset, is widely adopted for evaluations. Two types of accuracy are commonly used: overall accuracy (OA), which measures the overall performance of the algorithm, and mean accuracy (mAcc), which provides a class-specific measure of accuracy. OA is calculated as the ratio of the total number of correctly classified objects to the total number of objects in the dataset regardless of the class, while mAcc is calculated as the ratio of correctly classified objects to the total number of objects for each class, and then averaged to give an overall measure of performance.

\noindent\textbf{3D object detection} is the task of recognizing and localizing 3D objects in scene-level point clouds, 
aiming to estimate their precise positions and orientations. 3D bounding boxes are annotated as ground truth for training 3D detectors. Average precision (AP) is a commonly used evaluation metric, calculated based on precision and recall for a given set of objects and confidence thresholds. The metric compares ground-truth bounding boxes with predicted ones, and is calculated as the area under the precision-recall curve. The precision is calculated as the ratio of the number of correctly predicted objects to the total number of predicted objects, while the recall is calculated as the ratio of the number of correctly predicted objects to the total number of ground-truth objects.

\noindent\textbf{3D semantic segmentation} is the task of assigning semantic labels to each point in a 3D point cloud. Point-wise categorical annotations are collected as ground truth for this task.  IoU (Intersection over Union) and mean IoU (mIoU) are commonly used metrics for evaluations. IoU measures the overlap between the predicted and ground truth segmentations for a given class and is calculated as the ratio of the intersection to the union of the two sets. IoU is calculated for each class separately. mIoU is the mean of the IoU values across all classes and provides an overall measure of the segmentation model's performance.

\noindent\textbf{3D instance segmentation} is a task that involves assigning a unique instance ID to each object in a point cloud, thereby separating objects belonging to the same category and enabling more accurate object recognition and tracking. Point-wise instance annotations are needed to train models for this task. The mean average precision (mAP) is a popular evaluation metric used in 3D instance segmentation, computed as the mean of the average precision (AP, as used in 3D object detection) values across all classes. To calculate mAP, the precision-recall curve is computed for each class, and the area under the curve (AUC) is calculated. The AP is then calculated as the mean of the precision values at a set of predefined recall levels. Finally, the mAP is obtained as the mean of the AP values for all classes.

\noindent\textbf{Backbone.} A "backbone" is the essential and fundamental part of a neural network architecture that is responsible for extracting high-level features from input data. These features are then processed and analyzed by subsequent layers in the network. The backbone carries out most of the computation in a neural network and is crucial in determining its performance. To ensure fairness in comparing the performance of different label-efficient learning algorithms, it is important to use the same backbone implementation.

\vspace{-1.4cm}

\begin{IEEEbiography}[{\includegraphics[width=1in,height=1.25in,clip,keepaspectratio]{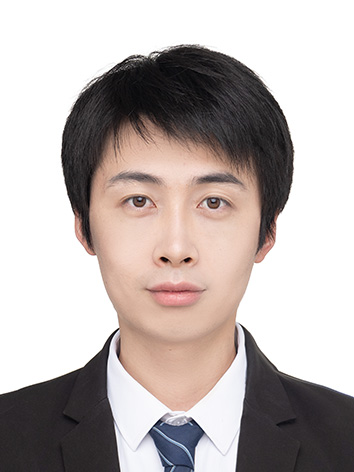}}]{Aoran Xiao} is a Research Fellow at the School of Computer Science and Engineering, Nanyang Technological University (NTU), Singapore. He obtained his Ph.D. in Computer Science and Engineering from NTU. Prior to this, he earned his B.Sc. and M.Sc. degrees in remote sensing from Wuhan University, China, in 2016 and 2019, respectively. His research interests include point cloud processing, computer vision, and remote sensing.
\end{IEEEbiography}

\vspace{-1.4cm}

\begin{IEEEbiography}[{\includegraphics[width=1in,height=1.25in,clip,keepaspectratio]{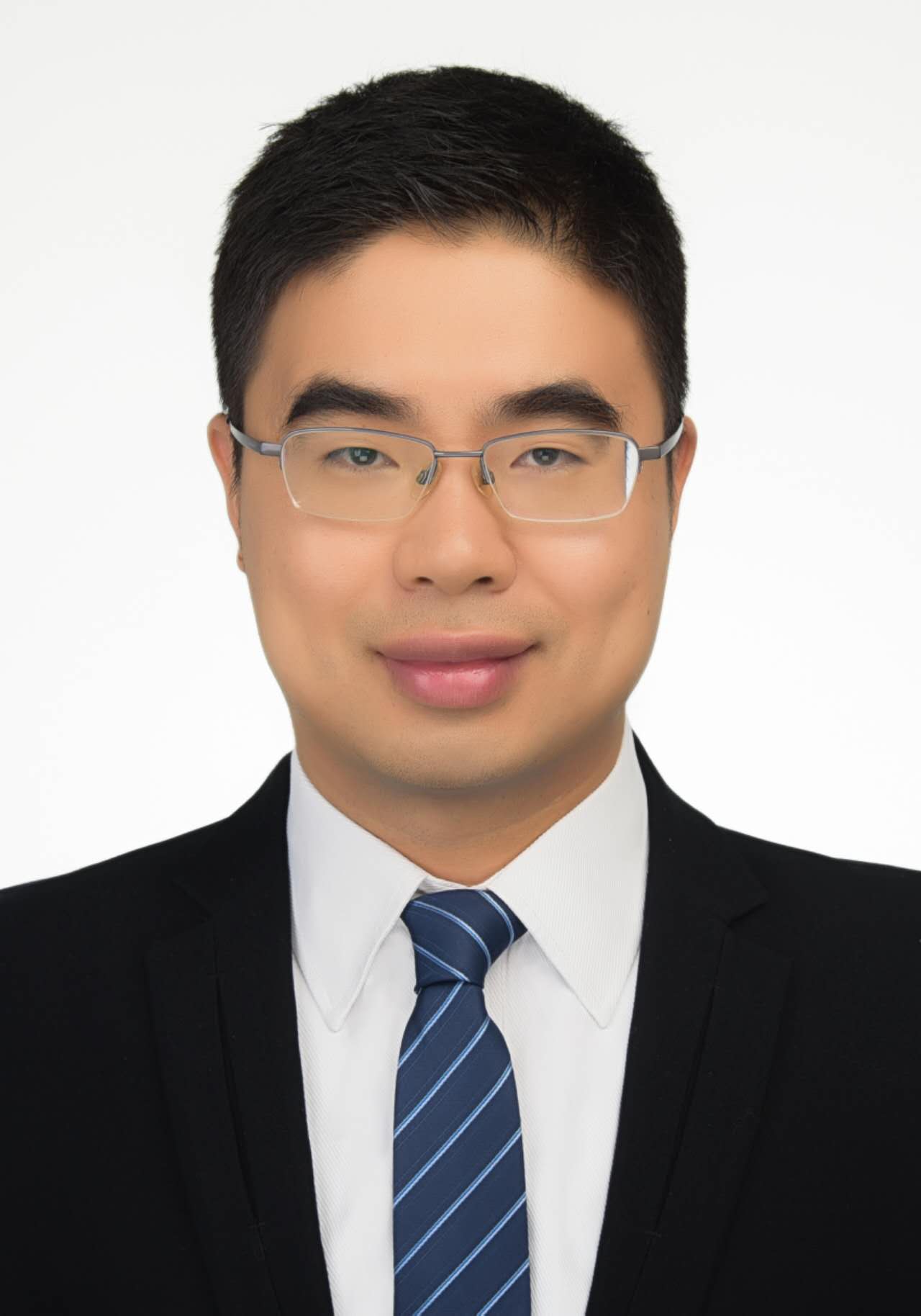}}]{Xiaoqin Zhang} is a senior member of the IEEE. He received the B.Sc. degree in electronic information science and technology from Central South University, China, in 2005, and the Ph.D. degree in pattern recognition and intelligent system from the National Laboratory of Pattern Recognition, Institute of Automation, Chinese Academy of Sciences, China, in 2010. He is currently a Professor with Wenzhou University, China. He has published more than 100 papers in international and national journals, and international conferences, including IEEE T-PAMI, IJCV, IEEE T-IP, IEEE T-NNLS, IEEE T-C, ICCV, CVPR, NIPS, IJCAI, AAAI, and among others. His research interests include in pattern recognition, computer vision, and machine learning.
\end{IEEEbiography}

\vspace{-1.4cm}

\begin{IEEEbiography}[{\includegraphics[width=1in,height=1.25in,clip,keepaspectratio]{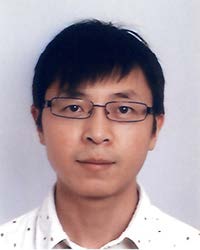}}]{Ling Shao} is a Distinguished Professor with the UCAS-Terminus AI Lab, University of Chinese Academy of Sciences, Beijing, China. He was the founding CEO and Chief Scientist of the Inception Institute of Artificial Intelligence, Abu Dhabi, UAE. His research interests include computer vision, deep learning, medical imaging and vision and language. He is a fellow of the IEEE, the IAPR, the BCS and the IET.
\end{IEEEbiography}

\vspace{-1.4cm}

\begin{IEEEbiography}[{\includegraphics[width=1in,height=1.25in,clip,keepaspectratio]{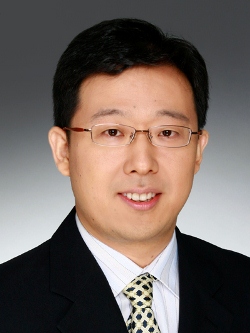}}]{Shijian Lu} is an Associate Professor with the School of Computer Science and Engineering at the Nanyang Technological University, Singapore. He received his PhD in electrical and computer engineering from the National University of Singapore. His major research interests include image and video analytics, visual intelligence, and machine learning.
\end{IEEEbiography}

\end{document}